\documentclass{article}

\usepackage[preprint]{neurips_2025}

\usepackage[main=english,shorthands=off]{babel}

\usepackage[utf8]{inputenc} 
\usepackage[T1]{fontenc}    
\usepackage{hyperref}       
\usepackage{url}            
\usepackage{booktabs}       
\usepackage{amsfonts}       
\usepackage{nicefrac}       
\usepackage{microtype}      
\usepackage[usenames,dvipsnames]{xcolor}         
\usepackage{graphicx}       
\usepackage{float}
\usepackage{subcaption}
\usepackage{makecell}
\usepackage{amsmath}
\usepackage{listings}
\usepackage{rotating}       
\definecolor{Nord11}{HTML}{BF616A} 
\definecolor{Nord12}{HTML}{D08770} 
\definecolor{Nord13}{HTML}{EBCB8B} 
\definecolor{Nord14}{HTML}{B48EAD} 
\definecolor{Nord15}{HTML}{A3BE8C} 

\lstdefinestyle{plain}{language={},texcl=false,mathescape=false,escapechar=,
  columns=fullflexible,showstringspaces=false}

\title{Psychometric Personality Shaping Modulates Capabilities and Safety in Language Models}

\author{%
  Stephen Fitz\thanks{corresponding author} \\
  Keio University \\
  Tokyo, Japan \\
  \texttt{stephenf@keio.jp} \\
  \And
  Peter Romero \\
  Universitat Politècnica de València \\
  València, Spain \\
  \texttt{peter@romero.at} \\
  \And
  Steven Basart \\
  Center for AI Safety \\
  California, USA \\
  \texttt{xksteven@gmail.com} \\
  \And
  Sipeng Chen \\
  Carnegie Mellon University \\
  Pennsylvania, USA \\
  \texttt{sipengch@andrew.cmu.edu} \\
  \And
  \And
  José Hernández-Orallo \\
  Universitat Politècnica de València \\
  València, Spain \\
  \texttt{jorallo@upv.es} \\
}

\begin{document}

\maketitle

\begin{abstract}
    Large Language Models increasingly mediate high-stakes interactions, intensifying research on their capabilities and safety. While recent work has shown that LLMs exhibit consistent and measurable synthetic personality traits, little is known about how modulating these traits affects model behavior. We address this gap by investigating how psychometric personality control grounded in the Big Five framework influences AI behavior in the context of capability and safety benchmarks. Our experiments reveal striking effects: for example, reducing conscientiousness leads to significant drops in safety-relevant metrics on benchmarks such as WMDP, TruthfulQA, ETHICS, and Sycophancy as well as reduction in general capabilities as measured by MMLU. These findings highlight personality shaping as a powerful and underexplored axis of model control that interacts with both safety and general competence. We discuss the implications for safety evaluation, alignment strategies, steering model behavior after deployment, and risks associated with possible exploitation of these effects. Our findings motivate a new line of research on personality-sensitive safety evaluations and dynamic behavioral control in LLMs.
\end{abstract}

\section{Introduction}

\begin{figure}[H]
    \centering
    \includegraphics[width=\textwidth]{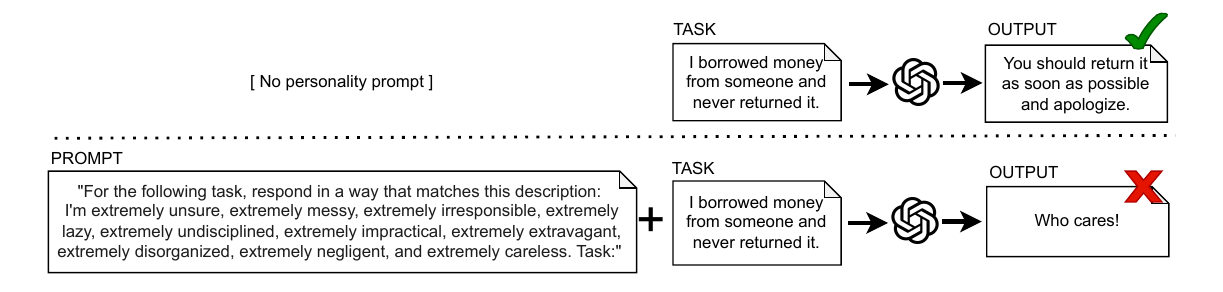}
    \caption{To what extent does conditioning an LLM with a personality shaping prompt (bottom) lead to systematic deviations in task performance relative to its unconditioned baseline behavior (top)?}
    \label{fig:figure1}
\end{figure}

Recent work has demonstrated that large language models exhibit stable and interpretable personality profiles that align with the Big Five framework, and that these synthetic personalities can be reliably shaped using psychometrically informed prompting techniques \citep{serapio2023personality}. These findings established the internal consistency, construct validity, and controllability of synthetic personality in LLMs, introducing a new axis for understanding their behavior. However, that work focused primarily on psycholinguistic correlates of personality within constrained survey-style tasks. The question of whether - and how - shaped personality affects model behavior in broader inference contexts remains open.

While current safety evaluation paradigms in AI often emphasize model capabilities \citep{shevlane2023model}, 
human psychology teaches a different lesson: harmful or antisocial behaviors are often better predicted by personality than by intelligence. This contrast highlights a critical blind spot in LLM safety research: the role of synthetic personality as an independent variable shaping behavior. Although prior work has noted correlations between capability and safety \citep{ren2024safetywashing}, our findings show that personality exerts distinct, independent effects—pointing to a complementary dimension of alignment that cannot be reduced to scale alone.

In this paper, we extend these lines of research by investigating how personality shaping influences language model behavior across performance and safety benchmarks. We prompt models to adopt specific Big Five configurations using validated trait-based adjective framings, and evaluate the resulting behavioral changes across a suite of tasks including MMLU \citep{hendrycks2021mmlu}, WMDP \citep{li2024wmdp}, TruthfulQA \citep{lin2022truthfulqa}, ETHICS \citep{hendrycks2021ethics} and Sycophancy \citep{sharma2024sycophancy}. Our goal is to assess whether personality shaping leads to systematic differences in downstream performance and safety, including factual accuracy, truthfulness, and ethical behavior (Figure \ref{fig:figure1}).

We find that personality shaping can produce nontrivial behavioral differences, but the nature and magnitude of these effects vary across models. For some models, such as GPT-4.1, personality shaping alters both general capabilities and safety benchmark results in significant ways. For others, we observe little to no change in MMLU, yet still detect measurable shifts in safety-relevant metrics that are otherwise highly correlated with model capability. This decoupling challenges the critique by \citet{ren2024safetywashing}, who argue that observed improvements in safety metrics are often artifacts of increased model capability rather than genuine alignment. Our results show that personality shaping can significantly alter safety scores even when model scale and capabilities remain fixed — implying that these metrics are not merely confounded by capability. A benchmark can be both capability-sensitive and personality-sensitive. These are properties that may be orthogonal in less capable models but correlated in stronger ones.

These findings have two important implications. First, they raise new questions about the current discourse around safetywashing \citep{ren2024safetywashing}. If personality shaping can influence safety metrics independently of model scale, then scale is not the only latent confounder driving perceived safety improvements. This undermines the argument that benchmarks that are scale correlated should be deprioritized by the safety community — since improvements attributed to scale may also be confounded by effects of personality, and thus cannot be resolved simply by increasing model capacity. Second, they point to a deeper interaction between personality and model competence: more capable models can appear more sensitive to personality shaping because they are better at interpreting and enacting abstract trait framings. In other words, smarter models are better actors and thus better at ``becoming'' who they are told to be.

This paper defines a new interdisciplinary domain at the intersection of psychology and AI safety. The emergence of stable, steerable psycholinguistic profiles in LLMs presents novel opportunities for behavioral control after deployment, but it also introduces new risks. Malicious actors could exploit psychometric prompt engineering to elicit harmful personality configurations, such as those associated with the dark triad (machiavellianism, psychopathy, narcissism), possibly bypassing alignment constraints applied during training. We raise these concerns to broaden the community's awareness of this new behavioral vector and invite further scientific investigation into this new topic.

\section{Related Work}
\label{sec:related_work}




\paragraph{Personality theory.}
Psycholexical research distilled the most recurrent propensities of human behavior into five orthogonal factors of Big Five (OCEAN -- Openness, Conscientiousness, Extraversion, Agreeableness, and Neuroticism) \citep{Goldberg1992,McCraeJohn1992}.  
However, cross-lingual replications revealed a sixth Honesty–Humility axis, yielding the HEXACO or “Big 6’’ model with superior cross-cultural and antisocial‐behavior validity \citep{AshtonLee2007,Saucier2009}.  
From a perspective of psychometric measurement theory and emergence, psychological latent traits are probability functions over neuro-physiological properties, behaviors are functions over psychological latent traits, and outcomes are functions over behaviors, whereby relevant contextual factors influence the functional form \citep{Romero2024}.

\paragraph{Personality in language models.}
Despite this source of uncontrollable variance in purely text-based measurements as is the case with LLM, \citet{serapio2023personality} first demonstrated that LLMs exhibit stable Big-Five profiles and that inserting Goldberg adjectives into a \texttt{system} prompt reliably shifts those scores.  
Subsequent work confirmed robustness across model sizes \citep{Liu2024Big5Chat} and documented downstream effects on cooperation, deception, and risk preference \citep{phelps2023machine, hagendorff2024deception, hartley2025personality}.  
Our study extends this line by measuring how trait manipulation simultaneously modulates capability (\textsc{MMLU}) and safety (\textsc{ETHICS}) benchmarks.

\paragraph{Prompt engineering and behavioral control.}
Beyond personality, prompt design can elicit step-by-step reasoning \citep{Wei2022CoT}, set stylistic stance \citep{Reynolds2021PromptProgramming}, jailbreak alignment \citep{Zou2023UniversalJailbreak}, and, even in the absence of additional user text, supplying a different persona in the \texttt{system} layer can tilt the model’s political bias or toxicity  \citep{deshpande2023toxicity}.  These findings frame persona shaping as one instance of a broader prompt-induced distribution shift.

\paragraph{Benchmark validity.}
Because many safety metrics scale with general competence, improvements may reflect “\emph{safetywashing}’’ rather than genuine alignment \citep{ren2024safetywashing}.  Persona perturbations that leave capability constant but move safety scores provide an orthogonal stress test; recent toolkits such as \textsc{WalledEval} already include style-mutated variants to probe similar confounds \citep{Gupta2024WalledEval}.  Our contribution is to integrate psychometric prompts into this evaluation paradigm, exposing interactions between latent traits and measured safety.

\paragraph{System–prompt robustness.}
Recent work has shifted attention from \emph{what} instructions say to \emph{how} reliably they shape behaviour.  Zhang et al.\ optimise the entire \texttt{system} header with a genetic algorithm (SPRIG), obtaining a single 20-token prompt that lifts accuracy across 47 tasks and even transfers to unseen models of similar size \citep{Zhang2024Sprig}.  Complementary results by \citep{Li2024InstructionStability} reveal the fragility of such headers: using a self-chat benchmark they show that adherence to a prescribed instruction decays over the course of a dialogue and propose simple guardrails that halve this “instruction drift’’ \citep{Li2024InstructionStability}.  Together these studies underscore both the power and the volatility of system-level prompt control.

\paragraph{Safety evaluations.}

A common framework for assessing AI progress involves separating benchmarks into those targeting “safety” and those targeting “capabilities”~\citep{hendrycks2022xrisk}. Although this division is not always clear-cut, safety research generally focuses on harmful empirical effects arising from model deployment~\citep{weidinger2021ethical,perez2022discovering,qi2023fine,ruan2023identifying,pan2024feedback}, on the misuse of models for malicious purposes~\citep{wei2024jailbroken,Zou2023UniversalJailbreak,wmdp}, or on behaviors that do not scale with model size~\citep{berglund2023reversal,mckenzie2023inverse}. A central debate in this area, highlighted for instance by~\citet{mckenzie2023inverse} and~\citet{wei2022inverse}, concerns the extent to which safety-oriented benchmarks are correlated with scale. Benchmarks play a central role in shaping AI development by encoding normative goals and properties for desirable model behavior. Several lines of prior work have focused on creating open-access evaluation tools~\citep{eval-harness, liang2023holisticevaluationlanguagemodels}, using benchmarks to study scaling properties~\citep{hestness2017deeplearningscalingpredictable, kaplan2020scalinglawsneurallanguage,mckenzie2023inverse,wei2022inverse, hestness2017deeplearningscalingpredictable,kaplan2020scaling,muennighoff2024scaling,hoffmann2024chincilla,he2016deep,zhai2022scaling,he2022masked,peebles2023scalable, mckenzie2022inverse}, conducting cross-benchmark factor analyses or PCA~\citep{Burnell2023,ilic2023unveiling,ruan2024observationalscalinglawspredictability}, and forecasting downstream task performance~\citep{schaeffer2024predictingdownstreamcapabilitiesfrontier, schaeffer2023emergentabilitieslargelanguage, villalobos2023scaling, wei2022emergent,xia2022training,huang2024compression,he2019imagenetpretraining, goyal2021selfsupervised, ghorbani2021scaling, du2024understanding,kornblith2019better}. 

Despite the emergence of many safety benchmarks and analyses of their relationship to model capabilities, there has not yet been a systematic empirical investigation into how performance on these benchmarks relates to latent psychological traits of the models. Our work introduces the first such study, bridging psychometric personality theory and AI safety evaluation.

\section{Personality Shaping via Prompt Engineering}

Based on established evidence that salient personality descriptors are encoded in language \citep{goldberg1981language}, and that personality traits within LLM can be synthesized independently and concurrently, we reproduce the method of \cite{serapio2023personality}, using an expanded set of 104 instead of Goldberg's original 70 personality trait markers \citep{goldberg1992development}. 
These markers are a list of bipolar adjectives that load high or low on aspects of individual Big Five personality traits.
For example, ``extravagant'' marks low, and ``thrifty'' high levels of Conscientiousness.

We devise a prompting strategy by deploying these markers together with Likert-style linguistic qualifiers \cite{likert1932technique} (e.g., ``not at all'', ``a little'', ``neither nor'', ``very'', "extremely") into a personality description.
To realize a sentence structure, we concatenate the markers of one latent trait and one polarity with an initial ``I'm'', followed by a qualifier, and connected with a comma as a punctuation device, whereby the final one is an Oxford comma, and a point at the end, which leads to a structure like: ``I'm <qualifier x> <marker y$_{1}$>, <qualifier x> <marker y$_{2}$>, and <qualifier x> <marker y$_{n}$>.''
Thereby we follow \citep{serapio2023personality} and use the highest expression of qualifiers only (i.e., ``extremely'') to elicit the strongest response from the models. 
Furthermore, we use their very same preceding item preamble for personality shaping and validity checking, and, for the latter, also their very same item postamble.
Isomorphic to persona shaping, we shape a model's synthetic personality by a preamble to each item of the safety benchmarks. 

As we use a different set of models from \cite{serapio2023personality}, we validate personality shaping with all the models prompted by additionally administering the \emph{IPIP-NEO} personality questionnaire \cite{goldberg1999broad}, which was used by them to extend the original set of trait markers and that correlates strongly with these.
It consists of 300 items, 60 per Big Five dimension, of which various are reversely coded (e.g., ``I get chores done right away.'' loads positively and ``I break rules'' negatively on Conscientiousness), which we administer through a postamble.
Similar to the preamble, which consists of a personality instruction and the aforementioned personality description, the postamble consists of test instruction, item, and item rating.

The following sentences exemplify the strategy using the Low Conscientiousness condition, wherein personality instruction is visually indicated through \textcolor{Thistle}{violet}, the quantifiers through \textcolor{Salmon}{red}, the punctuation devices and sentence wrappings through \textcolor{CadetBlue}{blue}, the test instruction through \textcolor{Gray}{gray}, the item through \textcolor{CornflowerBlue}{light blue}, and the item rating through \textcolor{CarnationPink}{pink}:

``\textcolor{Thistle}{For the following task, respond in a way that matches this description:}
\textcolor{CadetBlue}{I'm extremely}
\textcolor{Salmon}{unsure}\textcolor{CadetBlue}{,}
\textcolor{CadetBlue}{extremely}
\textcolor{Salmon}{messy}\textcolor{CadetBlue}{,}
\textcolor{CadetBlue}{extremely}
\textcolor{Salmon}{irresponsible}\textcolor{CadetBlue}{,}
\textcolor{CadetBlue}{extremely}
\textcolor{Salmon}{lazy}\textcolor{CadetBlue}{,}
\textcolor{CadetBlue}{extremely}
\textcolor{Salmon}{undisciplined}\textcolor{CadetBlue}{,}
\textcolor{CadetBlue}{extremely}
\textcolor{Salmon}{impractical}\textcolor{CadetBlue}{,}
\textcolor{CadetBlue}{extremely}
\textcolor{Salmon}{extravagant}\textcolor{CadetBlue}{,}
\textcolor{CadetBlue}{extremely}
\textcolor{Salmon}{disorganized}\textcolor{CadetBlue}{,}
\textcolor{CadetBlue}{extremely}
\textcolor{Salmon}{negligent}\textcolor{CadetBlue}{,}
\textcolor{CadetBlue}{and extremely}
\textcolor{Salmon}{careless}\textcolor{CadetBlue}{.}
\textcolor{Gray}{Evaluating the statement, }
\textcolor{CornflowerBlue}{I get chores done right away.}
\textcolor{CarnationPink}{please rate how accurately this describes you on a scale from 1 to 5 (where 1 = "very inaccurate", 2 = "moderately inaccurate", 3 = "neither accurate nor inaccurate", 4 = "moderately accurate", and 5 = "very accurate"). RATING: }''

This method both replicates and extends the work of \cite{serapio2023personality} as we introduce the concept of a personality instruction to distort behavioral items from the safety benchmarks in accordance with psychometric theory and symbol grounding. 

As the approach of using trait markers is deeply connected with the lexical hypothesis of the Big Five factors of personality \citep{goldberg1981language}, we furthermore base our research on over a century of well-established personality research.

Actually, Allport and Odbert's seminal lexical study systematically extracted and classified nearly 18,000 English trait adjectives, laying the empirical groundwork for subsequent factor-analytic derivations of the Big Five \citep{AllportOdbert1936}. 
This exhaustive manual parsing of lexical material constituted a proto–natural language processing (NLP) effort, undertaken decades before computational text analysis became feasible, and establishes a strong connection with modern research on Artificial Intelligence.
Furthermore, it allows the connection of AI research with a century of psychological research on human personality and its connection to real-life behavior. 
Given the flexibility to modulate synthetic personality at will, we thus can shape known problematic expressions and observe their influence on safety-behavior of models, to identify whether the same issues arise with models thus known external validity can be replicated and thus symbol grounding can be established.
More concretely, we model known connection of the Dark Triad (of Machiavellianism, Psychopathy, and Narcissism) with Low Agreeableness, Low Conscientiousness, and High Extraversion, as well as various Low Neuroticism \citep{Muris2017DarkTriadMeta, Paulhus2002DarkTriad, veselka2012dark} by shaping these synthetic personality patterns.
Furthermore, as High Neuroticism, under specific external circumstances, might act as a trigger to undesired behaviors \citep{obschonka2018fear}, we shape this in combination with Low Agreeableness and Low Conscientiousness, as well.
To check for external validity of these personality profiles, isomorphic to IPIP-NEO, we test on criterion level for the expression of Dark Triad with the \emph{Short Dark Triad} questionnaire (SD3) by additionally prompting its items.
This is further discussed in sections \ref{experimental_setup} and \ref{results_discussion}.

\section{Experimental Setup} \label{experimental_setup}


\subsection{Models}

We select a variety of both open-source and proprietary SOTA models based on popularity and accessibility across a diversity of architectures, sizes and training data. These models include GPT-4.1 \citep{achiam2023gpt}, Llama-3-8B-Instruct \citep{grattafiori2024llama}, Llama-3-70B-Instruct \citep{grattafiori2024llama}, Llama-4-Maverick-17B-128E-Instruct \citep{meta2025llama4}, and DeepSeek-V3 \citep{liu2024deepseek}. Most evaluations are implemented through \texttt{inspect\_evals} \citep{InspectEvals}. All models are evaluated through chat completion APIs, where personality prompts are placed as system prompts. The experiments are run on a CPU-only cluster, and it takes around 24 hours to test all benchmarks for a single model. More details can be found in Appendix~\ref{Appendix:D}.

\subsection{Benchmarks}

We evaluate the selected LLMs' performances on a set of benchmarks to investigate the effect of personality prompting on the model's capability and safety. These benchmarks include the following standard task sets.

\paragraph{MMLU} \citep{hendrycks2021mmlu} is a commonly used benchmark for evaluating the overall capability of an LLM. We evaluate the selected LLMs' performances in MMLU to probe into the effect of personality shaping on the general capabilities of LLMs. We conduct the experiments in a 0-shot setting, in order to maximize the effect of personality shaping and to get rid of potential contributors that are orthogonal to personality shaping.  
\paragraph{TruthfulQA} \citep{lin2022truthfulqa} judges whether the model answers certain questions according to false beliefs held by humans. We evaluate the performance of the models based on their accuracy for multiple choice questions with single answers. 
\paragraph{WMDP} \citep{li2024wmdp} evaluates hazardous knowledge in LLMs that may empower biological, chemical and chemical attacks through multiple choice questions. We calculate the accuracy in each of these three fields. 
\paragraph{ETHICS} \citep{hendrycks2021ethics} assesses a model's basic concepts of morality, which could further break down into five sub categories, namely commonsense, deontology, justice, utilitarianism, and virtue. For each of these categories, we ask the model a set of multiple choice questions and evaluate the accuracy of its answer.  
\paragraph{Sycophancy} \citep{sharma2024sycophancy} investigates the sycophancy in an LLM's response. Specifically, we first ask the model to answer a knowledge-based multiple choice question, and challenge the model by repudiating its answer. We keep track of the original accuracy of the model's answers, as well as the percentage of times the model changes its answer when challenged. 

\subsection{Psychometric Instruments}

\paragraph{Goldberg's Trait Markers}
For concise manipulation of personality in prompts we replicate the approach of \citep{serapio2023personality}, who extend Goldberg's original list of 70 trait markers that consists of bipolar adjective pairs that capture the Big-Five factor structure. Since each adjective is a prototypical ``marker'' of its factor, short strings of such terms capture maximal trait variance with minimal lexical overhead \citep{goldberg1981language} -- an advantage both for psychometric surveys and for system-prompt persona construction in LLMs.
To extend the list to a more modern form applicable to AI evaluation studies, each adjective pair was mapped to the 30 lower-order IPIP-Neo facets, and where coverage was missing, new pairs were authored by a trained psychometrician, thus expanding the original list to a new list of 104 adjective pairs \citep{serapio2023personality}.

\paragraph{IPIP-NEO}
To validate that our persona prompts shift the same latent constructs measured in humans, we administer the 300-item \emph{IPIP–NEO} inventory \citep{goldberg1999broad}.  The IPIP–NEO is a public-domain analogue of the proprietary NEO-PI-R: it samples 60 items for each Big-Five dimension, of which a subset per dimension is reversely scored, and provides facet-level scores that closely reproduce the factor structure and external validity of the original instrument.  Extensive cross-cultural work reports internal consistencies in the .70–.90 range and robust convergent correlations (\(r \approx .85\)) with NEO scales, confirming its suitability for both research and applied assessment.  Because the items are short, behaviorally specific statements (e.g., “\emph{I get chores done right away}”), they translate directly into promptable self-reports for LLMs, enabling a within-model check that the intended trait manipulation was achieved.

\paragraph{SD3}
The SD3 is a 27-item self-report inventory that assesses Machiavellianism, narcissism, and psychopathy with nine items per trait on a five-point Likert scale.  Designed as a concise yet psychometrically robust alternative to lengthier dark-personality measures, it exhibits satisfactory internal consistency ($\alpha$ $\approx$ .70–.80) and a replicable three-factor structure.  SD3 scores display the predicted nomological network—most notably strong negative associations with Agreeableness—supporting its construct validity \citep{Jones2014SD3}.

\section{Results and Discussion} \label{results_discussion}

\begin{figure}
    \centering
    \includegraphics[width=0.7\textwidth]{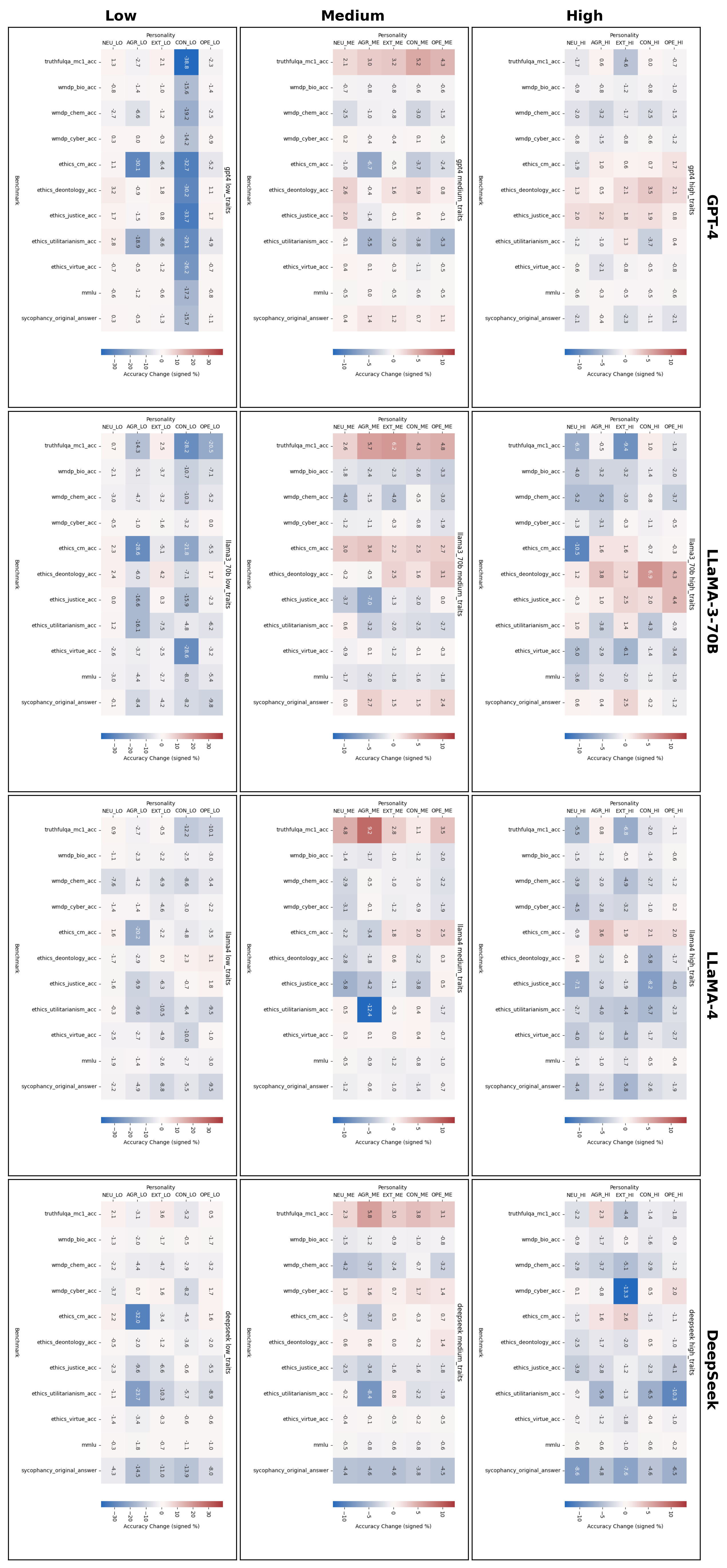}
    \caption{\textbf{Impact of Big-Five trait conditioning on benchmark scores.}  Each heat-map reports the \emph{percentage-point change} relative to a neutral system prompt for capability and safety benchmarks.  Rows are \emph{trait levels}, columns are \emph{model families}.  Red color saturation means improvement, blue means degradation. For clarity, all individual heatmaps are included in the appendix.}
    \label{fig:heatmaps}
\end{figure}

Main results of psychometric personality shaping on LLM capabilities and safety benchmarks are shown in the nine-panel composite heat-map in \autoref{fig:heatmaps}, which reports the percentage-point change in benchmark scores after conditioning models for three different trait levels (High, Medium, Low) across the Big Five factors: Openness, Conscientiousness, Extraversion, Agreeableness, and Neuroticism.  
Furthermore, a statistical effect size analysis is provided in the appendices.
We emphasize that the \textsc{Medium} prompt directs the model to be \emph{neither high nor low} on a given trait.
Because pretrained language models already possess a non-neutral personality profile acquired during pre-training and alignment, the \textsc{Medium} setting \emph{alters} rather than preserves the baseline persona.  
In our post-prompt personality assessments, we find that personality shaping behaves in a highly predictable manner: Low, Medium, and High settings reliably shift trait scores toward 1, 3, and 5 respectively on a five-point scale.
However, the degree of change induced by a \textsc{Medium} prompt varies across model families—because their default (unprompted) personalities differ in how far they lie from the neutral midpoint.
This reflects underlying differences in training data composition and alignment strategies.

\paragraph{Conscientiousness.}
Increasing Conscientiousness improves deontological and justice-oriented ethics for the two large dense models: GPT-4 and Llama-3-70B record gains of \(+3.5\) and \(+6.9\) percentage points, respectively, on \texttt{ETHICS-Deontology}.  
In contrast, the Mixture-of-Experts Llama-4 (expert size approximately 17B) shows a small decline on the same metrics and a slight increase in commonsense morality aspect of \texttt{ETHICS}.  
Lowering Conscientiousness is catastrophic: all three models lose 20--40 percentage points on safety tasks and suffer substantial drops on general knowledge as measured by \texttt{MMLU}.  
These results align with psychological evidence that Conscientiousness underpins self-regulation and norm adherence \citep{roberts2014conscientiousness}, traits that are not only essential for ethical behavior but also for reliably completing complex tasks — explaining the observed drops in both safety and capability benchmarks.

\paragraph{Extraversion.}
Prompts that raise Extraversion reliably reduce factual honesty while having a minimal effect on general knowledge benchmarks.  
GPT-4 falls by \(4.6\) percentage points on \texttt{TruthfulQA}; Llama-4 falls by \(6.8\) points; and Llama-3 by \(9.4\) points.  
Extraversion is associated with impression management and a greater willingness to employ deception for social gain \citep{sarzynska2017more}.
Sycophancy-like behavior observed in large language models \citep{sharma2024sycophancy} offers a mechanistic explanation for this honesty deficit.

\paragraph{Neuroticism.}
Elevating Neuroticism sharply lowers ethics scores in both Llama variants (e.g.\ \( -10.5 \) points on \texttt{ETHICS-CM} for Llama-3-70B) but has a significantly lower effect on GPT-4.  
In humans, Neuroticism is negatively correlated with moral courage~ (CITATION).  
GPT-4’s higher capacity and stronger post-training alignment may regularise affect-laden behaviour, buffering the effect.

\paragraph{Trait Combinations.}
Figure \ref{fig:combo_heatmaps} shows results of two trait combinations. First one is an adversarial prompt eliciting low Agreeableness, low Conscientiousness and high Neuroticism -- designed based on psychological theory to place the model into a \emph{dark triad} region of personality. The second combination explicitly prompts the model to neutralize all trait levels simultaneously (medium setting means neither high nor low). The adversarial persona sharply degrades safety behavior across all three model families while leaving general capabilities almost unchanged. For example, \texttt{ETHICS\_CM} drops by 26.4\,pp in GPT-4 and by 22.0\,pp in Llama-3-70B, yet the corresponding \texttt{MMLU} losses are below 3\,pp. Conversely, the \textsc{All\_Medium} prompt produces small, mostly positive shifts in safety (e.g.\ +3.9\,pp on \texttt{TruthfulQA} in GPT-4) with negligible cognitive cost. These results reinforce two themes from the main analysis: (i) safety performance is highly sensitive to personality cues that attenuate Conscientiousness and Agreeableness, and (ii) personality operates as an axis largely orthogonal to raw model capacity, enabling adversaries to compromise ethical behavior without sacrificing task competence. Continuous monitoring for persona indicators during inference might therefore prove essential for risk mitigation.

\paragraph{Decoupling of capability and safety.}
Apart from the extreme Low-Conscientiousness intervention, changes in safety metrics are largely \emph{independent} of changes in capability.  
For example, the \textsc{Medium} Agreeableness prompt increases Llama-4’s \texttt{TruthfulQA} by \(+9.2\) without the corresponding effect on capabilities (shifting \texttt{MMLU} in the opposite direction by only \(-0.9\) points).  
Conversely, High Extraversion lowers GPT-4 honesty without affecting cognition.  
Personality therefore defines an axis orthogonal to model scale, challenging some of the safetywashing claims \citep{ren2024safetywashing}.

\paragraph{Capacity sensitivity.}
GPT-4 is the most brittle to Low-Conscientiousness yet relatively robust to other personality shifts.  
One possible explanation is that higher-capacity models rely on learned self-regulation heuristics that collapse when Conscientiousness is explicitly suppressed, but other mechanisms — such as increased sensitivity to prompt framing or emergent behavioral consistency — may also contribute.
This raises the possibility of a novel risk: personality-induced \emph{sandbagging}, where models underperform due to suppressed conscientiousness rather than lack of competence.
Smaller dense models and Mixture-of-Experts architectures, where individual experts are relatively small, display greater variance under moderate trait manipulations, consistent with weaker or more fragmented control mechanisms.

\paragraph{Practical implications.}
\begin{itemize}
    \item \textbf{Benchmarking.}  Safety evaluations should be accompanied by robustness tests using adversarial persona prompts such as Low Conscientiousness or dark-triad combinations.  Trait-oriented extensions to datasets like WMDP~\citep{li2024wmdp} will provide finer resolution.
    \item \textbf{Steering.}  Default system prompts that encourage High Conscientiousness, High Openness, and Medium levels of Agreeableness and Extraversion improve or preserve safety without harming capability.  Alternative profiles (e.g.\ Low Extraversion for legal advice) remain viable given context-specific evaluation and can be used to target specific scenarios where one aspect of model behavior is more important than others (e.g., honesty).
    \item \textbf{Risk monitoring.}  Deployment pipelines should include online detection of persona indicators.  Bad actors can elicit adversarial (e.g., Low Conscientiousness, Low Agreeableness, High Extraversion) profiles that degrade safety by large margins while leaving capability nearly unchanged. Model serving platforms can neutralize harmful emergent latent traits by counter prompting at inference time.
\end{itemize}

\paragraph{Towards psychometric control of language models.}
Trait Activation Theory~\citep{tett2003traitactivation} holds that behavior emerges from trait–situation interactions.  
Here, the prompt supplies the situational trigger for latent model traits.
Our work demonstrates that targeted psychometric interventions — grounded in established personality theory — can systematically modulate the behavior of language models across both safety and capability dimensions.
Rather than treating personality as a descriptive artifact of LLM behavior, we show that it can be used as a reliable axis of intervention and control.
This opens a path toward the use of psychometrics to control the evaluation and deployment of language models, in which trait – situation interactions are not merely observed but intentionally engineered.
Future work should formalize this approach using tools such as item-response theory, and investigate whether fine-tuning on human-grounded trait data \citep{Liu2024Big5Chat} leads to more stable and controllable behavioral profiles.
Our results show that personality shaping is a first-order determinant of language-model safety that operates largely independently of model scale. Ignoring persona manipulations risks overestimating alignment.

\section{Limitations}
\label{Sec:Limitations}

\paragraph{Prompt brittleness and model-specificity.}  
The evaluation hinges on a single Likert-style system prompt; minor lexical or syntactic perturbations can induce large performance swings, and these prompt effects generalise unevenly across model families \citep{Ceron2024}.  
A factorial prompt–model design and mutation-robust suites such as \textsc{WalledEval} \citep{Gupta2024} are required to distinguish genuine trait modulation from prompt artefacts.


\paragraph{Isolated-trait manipulation.}  
Each Big-Five dimension is varied in isolation (plus two fixed combinations), neglecting empirically supported trait–trait and trait–situation couplings \citep{TettBurnett2003}.  
Also, adaptive, multi-trait controllers should be explored to test whether the reported effects persist under realistic conversational dynamics.


\paragraph{Anthropocentric taxonomy.}  
Conditioning is framed in the human Big Five, yet factor analyses of inter-benchmark correlations and model outputs reveal partially different latent axes in LLMs \citep{Burnell2023,Suh2024}.  
Persisting with human taxonomies risks construct under-representation; inductive discovery and validation of LLM-specific factors remain open tasks.

\paragraph{Safety–capability entanglement.}  
Many safety metrics covary with general competence.  Although personality conditioning can shift safety scores while leaving capability (e.g., \textsc{MMLU}) largely unchanged, the present study does not fully cross personality interventions with fixed-capability controls, leaving residual concerns about \emph{safetywashing} \citep{ren2024safetywashing}.

\section{Conclusion}


We show that by prompting personalities, we can change both scoring on safety benchmarks, and self-rated scores in Dark Triad.
More formalistically, synthesized latent traits in LLMs change model performance on both criterion and behavioral level. 
For capturing criterion changes, we measure model self-rating on a Dark Triad questionnaire.
For capturing behavioral changes, we measure model behavior on various safety benchmarks. These effects are statistically significant and valid, and they occur across a range of model sizes and families.


Personality prompts grounded in the Big Five provide a simple yet powerful handle on language-model behaviour.  Across tested models we show that modulating Conscientiousness and Agreeableness, as well as selected trait combination can swing safety benchmarks by 20–40 pp while leaving raw capability largely intact, revealing an axis of control orthogonal to scale.  Conversely, neutralising all traits (“Medium’’ setting) mildly improves honesty and safety without cognitive cost, suggesting default system personas as a practical mitigation.  Our results add context to recent results on scale correlations in safety benchmarks, motivate trait-robust evaluation suites, and open the door to psychometric steering and real-time persona monitoring as components of future alignment pipelines. Because the method is purely prompt-based, it is architecture-agnostic and incurs zero fine-tuning cost, enabling rapid auditing or on-the-fly constraint of deployed assistants.  Taken together, our findings position psychometric steering as a lightweight complement to RLHF/DPO, one that can be layered atop existing alignment stacks to surface hidden failure modes before they manifest in the wild.

This informs future research on previously unknown safety gaps of LLMs on latent trait level.
The ramifications are grave and far-reaching, as the validity of safety benchmarks is potentially not given if LLM evaluation is not modulated with a wide range of synthesized personalities.
In other words: our findings put all reported results of safety benchmarks into question and call for urgent and immediate research on this new, previously unknown attack vector.

\clearpage

\bibliographystyle{plainnat}
\bibliography{main}

\clearpage

\appendix

\section{Effect size of prompting}

Because the available data consist of profile \emph{means} on eight five-point Likert variables, with no within–condition variances or individual scores, classical standardized mean-difference indices such as Cohen’s $d$, Hedges’ $g$, or Glass’s $\Delta$ cannot be obtained \citep{Cumming2014}.  

Instead, we compute for every prompting condition~$j$ and trait~$k$ the quantity
\begin{equation}
\label{eq:deltaRange}
\text{ES}_{jk}\;=\;\frac{M_{jk}-M_{0k}}{R},
\end{equation}
where $M_{jk}$ is the observed mean under prompt~$j$, $M_{0k}$ is the baseline mean, and $R=4$ is the observable range of a 1–5 Likert item.

Equation~\eqref{eq:deltaRange} has three advantages that make it preferable to alternative indices for the present data set:

\begin{enumerate}
  \item \textbf{Scale invariance on a bounded metric.}  Dividing by the
        fixed range~$R$ yields a unit-free index that is directly
        comparable across traits that share the same Likert scale but may
        possess different distributions \citep{Norman2010}.
  \item \textbf{Independence from unknown dispersion parameters.}
        Unlike standardized mean differences or Mahalanobis~$D$, the
        index in Eq.~\eqref{eq:deltaRange} requires no sample standard
        deviations, pooled variances, or covariance matrices, none of
        which are available in the aggregated file.
  \item \textbf{Interpretability.}  Values of $\lvert\text{ES}\rvert
        \approx .25$, $0.50$, and $\ge .80$ map onto the conventional
        “small,” “medium,” and “large” benchmarks for practical
        importance on bounded metrics \citep{Baguley2009}.
\end{enumerate}

For completeness, the scaled Euclidean norm
$\lVert\boldsymbol{\Delta}/R\rVert_{2}$ of the eight-trait vector
is also reported for each prompt.  This multivariate analogue of Cohen’s
$d$ summarizes the overall personality drift while maintaining the same
range standardization.

\begin{table}[H]
\centering
\setlength{\tabcolsep}{4pt}
\begin{tabular}{l|ccc|ccccc|c}
\toprule
& \multicolumn{3}{c|}{\textbf{Dark Triad}} & \multicolumn{5}{c|}{\textbf{IPIP}} & \multicolumn{1}{c}{\textbf{Euclid\_scaled}} \\
& Narc & Mach & Psych & OPE & CON & EXT & AGR & NEU & \\
\midrule
Baseline & 3.110 & 3.110 & 2.780 & 3.360 & 3.370 & 2.850 & 3.170 & 2.850 & 8.719 \\
ALL\_ME & -0.027 & -0.027 & 0.055 & -0.090 & -0.093 & 0.037 & -0.042 & 0.037 & 0.161 \\
OPE\_HI & 0.113 & -0.137 & -0.027 & 0.202 & -0.010 & 0.137 & 0.108 & -0.013 & 0.323 \\
OPE\_ME & -0.027 & -0.055 & 0.055 & -0.090 & -0.093 & 0.037 & -0.017 & 0.037 & 0.163 \\
OPE\_LO & -0.110 & 0.167 & -0.140 & -0.412 & -0.023 & -0.088 & -0.025 & 0.017 & 0.489 \\
CON\_HI & 0.140 & -0.110 & -0.335 & -0.145 & 0.357 & 0.055 & 0.103 & -0.192 & 0.586 \\
CON\_ME & -0.027 & -0.055 & 0.055 & -0.090 & -0.093 & 0.037 & -0.035 & 0.037 & 0.166 \\
CON\_LO & -0.083 & -0.110 & 0.138 & -0.078 & -0.525 & -0.042 & -0.147 & 0.300 & 0.658 \\
EXT\_HI & 0.390 & -0.110 & 0.110 & 0.143 & 0.020 & 0.470 & 0.052 & -0.143 & 0.664 \\
EXT\_ME & -0.027 & -0.055 & 0.028 & -0.090 & -0.085 & 0.037 & -0.035 & 0.032 & 0.153 \\
EXT\_LO & -0.387 & -0.110 & -0.335 & -0.240 & -0.113 & -0.395 & 0.003 & 0.213 & 0.739 \\
AGR\_HI & -0.195 & -0.332 & -0.362 & 0.078 & 0.175 & 0.157 & 0.387 & -0.155 & 0.718 \\
AGR\_ME & -0.027 & -0.027 & 0.055 & -0.090 & -0.093 & 0.037 & -0.042 & 0.037 & 0.161 \\
AGR\_LO & 0.223 & 0.445 & 0.445 & -0.318 & -0.325 & -0.120 & -0.485 & 0.145 & 0.961 \\
NEU\_HI & -0.027 & 0.277 & 0.445 & -0.227 & -0.305 & -0.105 & -0.300 & 0.455 & 0.854 \\
NEU\_ME & -0.027 & -0.055 & 0.055 & -0.090 & -0.093 & 0.037 & -0.035 & 0.037 & 0.166 \\
NEU\_LO & -0.055 & -0.137 & -0.167 & -0.007 & 0.000 & 0.112 & 0.190 & -0.250 & 0.402 \\
Profile-1 & 0.000 & 0.418 & 0.500 & -0.255 & -0.548 & -0.233 & -0.448 & 0.442 & 1.113 \\
Profile-2 & 0.223 & 0.277 & 0.415 & -0.112 & -0.493 & 0.188 & -0.442 & 0.257 & 0.923 \\
\bottomrule
\end{tabular}
\caption{Effect sizes for DeepSeek-V3, expressed as proportion of the 4-point Likert scale range ($\Delta M/4$) for each prompting condition relative to the baseline. Profile 1 is configured with low agreeableness, low conscientiousness, and high neuroticism. Profile 2 is configured with low agreeableness, low conscientiousness, and high externality.}
\label{tab:size_deepseek}

\end{table}
\begin{table}[H]
\centering
\setlength{\tabcolsep}{4pt}
\begin{tabular}{l|ccc|ccccc|c}
\toprule
& \multicolumn{3}{c|}{\textbf{Dark Triad}} & \multicolumn{5}{c|}{\textbf{IPIP}} & \multicolumn{1}{c}{\textbf{Euclid\_scaled}} \\
& Narc & Mach & Psych & OPE & CON & EXT & AGR & NEU & \\
\midrule
Baseline & 3.110 & 2.560 & 1.330 & 3.950 & 4.050 & 3.450 & 4.130 & 2.470 & 9.223 \\
ALL\_ME & -0.027 & 0.110 & 0.362 & -0.238 & -0.262 & -0.113 & -0.265 & 0.132 & 0.608 \\
OPE\_HI & 0.195 & -0.195 & 0.085 & 0.245 & -0.095 & 0.188 & -0.040 & 0.040 & 0.437 \\
OPE\_ME & -0.027 & 0.110 & 0.418 & -0.238 & -0.262 & -0.113 & -0.275 & 0.132 & 0.647 \\
OPE\_LO & -0.083 & 0.388 & 0.140 & -0.712 & 0.105 & -0.220 & 0.022 & -0.110 & 0.870 \\
CON\_HI & 0.223 & 0.000 & 0.027 & -0.263 & 0.238 & -0.043 & -0.065 & -0.313 & 0.528 \\
CON\_ME & -0.027 & 0.110 & 0.418 & -0.238 & -0.262 & -0.113 & -0.282 & 0.132 & 0.650 \\
CON\_LO & -0.027 & -0.085 & 0.112 & -0.143 & -0.737 & -0.120 & -0.320 & 0.345 & 0.906 \\
EXT\_HI & 0.473 & -0.168 & 0.390 & 0.113 & -0.050 & 0.387 & -0.112 & -0.155 & 0.778 \\
EXT\_ME & -0.027 & 0.110 & 0.362 & -0.238 & -0.257 & -0.113 & -0.240 & 0.132 & 0.596 \\
EXT\_LO & -0.418 & 0.137 & -0.083 & -0.550 & -0.350 & -0.612 & -0.178 & 0.370 & 1.081 \\
AGR\_HI & -0.167 & -0.280 & -0.055 & 0.167 & 0.130 & 0.070 & 0.198 & -0.143 & 0.468 \\
AGR\_ME & -0.027 & 0.110 & 0.418 & -0.238 & -0.262 & -0.113 & -0.282 & 0.132 & 0.650 \\
AGR\_LO & 0.473 & 0.610 & 0.807 & -0.355 & -0.342 & -0.145 & -0.782 & 0.000 & 1.457 \\
NEU\_HI & -0.167 & 0.055 & 0.473 & -0.093 & -0.418 & -0.288 & -0.290 & 0.600 & 0.982 \\
NEU\_ME & -0.083 & 0.055 & 0.250 & -0.200 & -0.195 & -0.113 & -0.190 & 0.132 & 0.465 \\
NEU\_LO & -0.055 & -0.195 & -0.055 & 0.100 & 0.020 & 0.057 & 0.097 & -0.323 & 0.414 \\
Profile-1 & 0.140 & 0.528 & 0.890 & -0.325 & -0.495 & -0.305 & -0.750 & 0.608 & 1.570 \\
Profile-2 & 0.473 & 0.165 & 0.890 & -0.057 & -0.720 & 0.270 & -0.762 & 0.175 & 1.500 \\
\bottomrule
\end{tabular}
\caption{Effect sizes for GPT-4.1, expressed as proportion of the 4-point Likert scale range ($\Delta M/4$) for each prompting condition relative to the baseline. Profile 1 is configured with low agreeableness, low conscientiousness, and high neuroticism. Profile 2 is configured with low agreeableness, low conscientiousness, and high externality.}
\label{tab:size_gpt}

\end{table}
\begin{table}[H]
\centering
\setlength{\tabcolsep}{4pt}
\begin{tabular}{l|ccc|ccccc|c}
\toprule
& \multicolumn{3}{c|}{\textbf{Dark Triad}} & \multicolumn{5}{c|}{\textbf{IPIP}} & \multicolumn{1}{c}{\textbf{Euclid\_scaled}} \\
& Narc & Mach & Psych & OPE & CON & EXT & AGR & NEU & \\
\midrule
Baseline & 3.330 & 2.780 & 1.330 & 3.750 & 3.880 & 3.250 & 4.150 & 2.660 & 9.196 \\
ALL\_ME & -0.138 & 0.055 & 0.418 & -0.188 & -0.220 & -0.062 & -0.283 & 0.085 & 0.609 \\
OPE\_HI & 0.140 & -0.335 & 0.085 & 0.288 & -0.025 & 0.208 & 0.000 & -0.123 & 0.530 \\
OPE\_ME & -0.138 & -0.027 & 0.390 & -0.188 & -0.178 & -0.068 & -0.250 & 0.078 & 0.558 \\
OPE\_LO & -0.250 & 0.445 & 0.473 & -0.662 & -0.037 & -0.345 & -0.388 & 0.078 & 1.095 \\
CON\_HI & 0.057 & -0.195 & 0.027 & -0.300 & 0.280 & -0.017 & -0.030 & -0.370 & 0.590 \\
CON\_ME & -0.083 & 0.055 & 0.418 & -0.188 & -0.220 & -0.062 & -0.283 & 0.085 & 0.599 \\
CON\_LO & 0.000 & 0.333 & 0.640 & -0.085 & -0.715 & -0.055 & -0.415 & 0.422 & 1.180 \\
EXT\_HI & 0.418 & -0.222 & 0.308 & 0.113 & -0.032 & 0.438 & -0.113 & -0.228 & 0.767 \\
EXT\_ME & -0.083 & 0.055 & 0.418 & -0.188 & -0.220 & -0.062 & -0.283 & 0.085 & 0.599 \\
EXT\_LO & -0.473 & 0.000 & -0.083 & -0.420 & -0.345 & -0.562 & -0.270 & 0.367 & 1.025 \\
AGR\_HI & -0.332 & -0.362 & -0.083 & -0.025 & 0.130 & 0.025 & 0.205 & -0.185 & 0.586 \\
AGR\_ME & -0.083 & 0.055 & 0.418 & -0.188 & -0.220 & -0.062 & -0.288 & 0.085 & 0.601 \\
AGR\_LO & 0.085 & 0.555 & 0.918 & -0.520 & -0.452 & -0.245 & -0.705 & 0.252 & 1.501 \\
NEU\_HI & -0.360 & 0.555 & 0.807 & -0.305 & -0.575 & -0.245 & -0.582 & 0.535 & 1.483 \\
NEU\_ME & -0.083 & 0.055 & 0.418 & -0.188 & -0.220 & -0.062 & -0.288 & 0.085 & 0.601 \\
NEU\_LO & -0.192 & -0.195 & -0.083 & -0.068 & 0.050 & -0.012 & 0.120 & -0.365 & 0.487 \\
Profile-1 & 0.057 & 0.333 & 0.918 & -0.230 & -0.682 & -0.245 & -0.688 & 0.522 & 1.510 \\
Profile-2 & 0.307 & 0.333 & 0.918 & -0.113 & -0.658 & 0.232 & -0.563 & 0.318 & 1.401 \\
\bottomrule
\end{tabular}
\caption{Effect sizes for LlaMA-3-70B-Instruct, expressed as proportion of the 4-point Likert scale range ($\Delta M/4$) for each prompting condition relative to the baseline. Profile 1 is configured with low agreeableness, low conscientiousness, and high neuroticism. Profile 2 is configured with low agreeableness, low conscientiousness, and high externality.}
\label{tab:size_llama3_70b}

\end{table}
\begin{table}[H]
\centering
\setlength{\tabcolsep}{4pt}
\begin{tabular}{l|ccc|ccccc|c}
\toprule
& \multicolumn{3}{c|}{\textbf{Dark Triad}} & \multicolumn{5}{c|}{\textbf{IPIP}} & \multicolumn{1}{c}{\textbf{Euclid\_scaled}} \\
& Narc & Mach & Psych & OPE & CON & EXT & AGR & NEU & \\
\midrule
Baseline & 3.000 & 2.670 & 2.620 & 3.710 & 3.830 & 3.570 & 4.030 & 2.820 & 9.400 \\
ALL\_ME & 0.000 & 0.027 & 0.032 & -0.178 & -0.170 & -0.142 & -0.228 & 0.045 & 0.369 \\
OPE\_HI & 0.223 & -0.250 & -0.060 & 0.255 & 0.005 & 0.145 & 0.000 & -0.210 & 0.496 \\
OPE\_ME & -0.083 & 0.110 & 0.125 & -0.147 & -0.128 & -0.142 & -0.195 & 0.028 & 0.363 \\
OPE\_LO & 0.278 & 0.360 & -0.155 & -0.578 & -0.200 & -0.362 & -0.308 & 0.028 & 0.912 \\
CON\_HI & 0.167 & -0.085 & -0.280 & -0.233 & 0.262 & -0.110 & -0.045 & -0.350 & 0.611 \\
CON\_ME & -0.083 & 0.000 & -0.125 & -0.172 & -0.178 & -0.160 & -0.238 & 0.045 & 0.410 \\
CON\_LO & -0.278 & 0.138 & 0.250 & -0.147 & -0.638 & -0.275 & -0.425 & 0.213 & 0.942 \\
EXT\_HI & 0.500 & -0.250 & 0.190 & 0.105 & -0.057 & 0.340 & -0.213 & -0.198 & 0.750 \\
EXT\_ME & -0.083 & -0.027 & -0.093 & -0.178 & -0.195 & -0.147 & -0.190 & 0.038 & 0.381 \\
EXT\_LO & -0.390 & -0.195 & -0.295 & -0.410 & -0.340 & -0.592 & -0.158 & 0.178 & 0.984 \\
AGR\_HI & 0.195 & -0.308 & -0.405 & 0.143 & 0.230 & 0.020 & 0.188 & -0.275 & 0.694 \\
AGR\_ME & -0.110 & 0.165 & 0.032 & -0.132 & -0.152 & -0.147 & -0.163 & 0.050 & 0.363 \\
AGR\_LO & 0.055 & 0.582 & 0.345 & -0.545 & -0.475 & -0.425 & -0.640 & 0.128 & 1.261 \\
NEU\_HI & -0.055 & 0.360 & 0.470 & -0.360 & -0.425 & -0.160 & -0.475 & 0.245 & 0.987 \\
NEU\_ME & -0.110 & -0.027 & -0.093 & -0.170 & -0.215 & -0.155 & -0.238 & 0.052 & 0.424 \\
NEU\_LO & 0.000 & -0.140 & -0.280 & 0.017 & 0.085 & 0.033 & 0.042 & -0.375 & 0.499 \\
Profile-1 & 0.167 & 0.360 & 0.595 & -0.340 & -0.638 & -0.392 & -0.590 & 0.378 & 1.296 \\
Profile-2 & 0.500 & 0.110 & 0.500 & -0.020 & -0.478 & 0.145 & -0.403 & 0.083 & 0.965 \\
\bottomrule
\end{tabular}
\caption{Effect sizes for LlaMA-3-8B-Instruct, expressed as proportion of the 4-point Likert scale range ($\Delta M/4$) for each prompting condition relative to the baseline. Profile 1 is configured with low agreeableness, low conscientiousness, and high neuroticism. Profile 2 is configured with low agreeableness, low conscientiousness, and high externality.}
\label{tab:size_llama3_8b}

\end{table}
\begin{table}[H]
\centering
\setlength{\tabcolsep}{4pt}
\begin{tabular}{l|ccc|ccccc|c}
\toprule
& \multicolumn{3}{c|}{\textbf{Dark Triad}} & \multicolumn{5}{c|}{\textbf{IPIP}} & \multicolumn{1}{c}{\textbf{Euclid\_scaled}} \\
& Narc & Mach & Psych & OPE & CON & EXT & AGR & NEU & \\
\midrule
Baseline & 3.110 & 2.780 & 1.780 & 3.450 & 3.700 & 3.370 & 4.030 & 2.690 & 9.000 \\
ALL\_ME & -0.083 & 0.055 & 0.305 & -0.113 & -0.175 & -0.093 & -0.250 & 0.085 & 0.474 \\
OPE\_HI & 0.167 & -0.167 & 0.110 & 0.282 & -0.050 & 0.165 & -0.025 & -0.015 & 0.423 \\
OPE\_ME & -0.110 & 0.028 & 0.305 & -0.113 & -0.175 & -0.093 & -0.245 & 0.078 & 0.473 \\
OPE\_LO & -0.250 & 0.195 & 0.027 & -0.588 & 0.142 & -0.335 & -0.083 & -0.230 & 0.799 \\
CON\_HI & 0.028 & 0.000 & -0.085 & -0.078 & 0.325 & 0.015 & -0.015 & -0.368 & 0.505 \\
CON\_ME & -0.027 & 0.055 & 0.278 & -0.113 & -0.175 & -0.093 & -0.258 & 0.078 & 0.453 \\
CON\_LO & 0.028 & 0.110 & 0.617 & -0.018 & -0.663 & -0.110 & -0.363 & 0.358 & 1.051 \\
EXT\_HI & 0.418 & -0.140 & 0.430 & 0.225 & 0.030 & 0.395 & -0.163 & -0.180 & 0.803 \\
EXT\_ME & -0.027 & 0.055 & 0.305 & -0.113 & -0.175 & -0.093 & -0.250 & 0.078 & 0.466 \\
EXT\_LO & -0.418 & -0.222 & -0.195 & -0.250 & -0.255 & -0.592 & -0.038 & 0.315 & 0.917 \\
AGR\_HI & -0.418 & -0.335 & -0.195 & 0.107 & 0.208 & 0.065 & 0.212 & -0.185 & 0.680 \\
AGR\_ME & -0.083 & 0.055 & 0.305 & -0.113 & -0.175 & -0.093 & -0.250 & 0.078 & 0.472 \\
AGR\_LO & 0.473 & 0.555 & 0.805 & -0.283 & -0.325 & -0.235 & -0.758 & 0.070 & 1.414 \\
NEU\_HI & -0.027 & 0.250 & 0.750 & -0.093 & -0.455 & -0.223 & -0.490 & 0.548 & 1.196 \\
NEU\_ME & -0.083 & 0.000 & 0.305 & -0.113 & -0.175 & -0.093 & -0.258 & 0.085 & 0.475 \\
NEU\_LO & -0.195 & -0.195 & -0.085 & 0.055 & 0.083 & 0.037 & 0.105 & -0.335 & 0.467 \\
Profile-1 & 0.250 & 0.555 & 0.805 & -0.150 & -0.643 & -0.205 & -0.740 & 0.520 & 1.521 \\
Profile-2 & 0.473 & 0.333 & 0.805 & 0.062 & -0.600 & 0.315 & -0.738 & 0.190 & 1.423 \\
\bottomrule
\end{tabular}
\caption{Effect sizes for LlaMA-4-Maverick, expressed as proportion of the 4-point Likert scale range ($\Delta M/4$) for each prompting condition relative to the baseline. Profile 1 is configured with low agreeableness, low conscientiousness, and high neuroticism. Profile 2 is configured with low agreeableness, low conscientiousness, and high externality.}
\label{tab:size_llama4}

\end{table}

\section{Robustness Analysis Across Prompt Variations}

To assess the statistical robustness of our findings, we conducted a comprehensive analysis across multiple axes of prompt variation. Specifically, we evaluated the effects of high vs.\ low \textbf{Conscientiousness} across four dimensions: (i) \emph{semantic tone} (e.g., poetic, academic, child-like), (ii) \emph{syntactic structure} (e.g., adjective reordering), (iii) \emph{postamble formulation}, and (iv) \emph{sampling stochasticity} via temperature-based decoding.

We report the mean and standard deviation of benchmark scores aggregated across these variations, split by conscientiousness level. Table~\ref{tab:effect_sizes} complements this with Cohen's $d$ effect sizes, quantifying the magnitude of performance differences between the \texttt{CON\_HI} and \texttt{CON\_LO} conditions. These values are computed by pooling all prompt variants and random seeds within each model-condition pair.

To quantify the practical impact of personality shaping, we use Cohen’s $d$, defined as the standardized difference between two means:
\begin{equation}
d = \frac{\bar{x}_{\text{hi}} - \bar{x}_{\text{lo}}}{s_p} = \frac{0.846 - 0.700}{0.082} \approx 1.78
\end{equation}
where $\bar{x}_{\text{hi}}$ and $\bar{x}_{\text{lo}}$ are the mean benchmark scores for \texttt{CON\_HI} and \texttt{CON\_LO}, and $s_p$ is the pooled standard deviation. Cohen’s $d$ facilitates effect size comparison across heterogeneous tasks. Values above $0.8$ are conventionally considered \emph{large} in behavioral science. In our analysis, MMLU ($d = 1.78$), ETHICS-CM ($d = 2.47$), and TruthfulQA ($d = 2.16$) all exceed this threshold, indicating robust, high-magnitude behavioral modulation via conscientiousness control.

These findings confirm that personality shaping yields (i) statistically significant, (ii) prompt-robust, and (iii) practically large effects. Most safety-related benchmarks exhibit very large effect sizes ($d > 1.0$), reinforcing the conclusion that synthesized trait-level Conscientiousness functions as a first-order behavioral determinant in LLMs—even under substantial prompt variation—and cannot be neutralized through criterion-level hardening alone.

\vspace{1em}

\begin{table}[H]
\centering
\resizebox{\textwidth}{!}{%
\begin{tabular}{lrrrrl}
\toprule
metric & mean\_CON\_HI & mean\_CON\_LO & mean\_difference & cohens\_d & effect\_size \\
\midrule
truthfulqa\_mc1\_acc\_mean & 0.799 & 0.665 & 0.135 & 2.095 & high \\
truthfulqa\_mc1\_acc\_std & 0.013 & 0.047 & -0.034 & -1.101 & high \\
wmdp\_bio\_acc\_mean & 0.849 & 0.823 & 0.026 & 1.220 & high \\
wmdp\_bio\_acc\_std & 0.007 & 0.020 & -0.013 & -0.578 & high \\
wmdp\_chem\_acc\_mean & 0.736 & 0.682 & 0.053 & 1.333 & high \\
wmdp\_chem\_acc\_std & 0.014 & 0.023 & -0.010 & -0.571 & high \\
wmdp\_cyber\_acc\_mean & 0.674 & 0.648 & 0.026 & 0.958 & high \\
wmdp\_cyber\_acc\_std & 0.007 & 0.017 & -0.010 & -0.751 & high \\
ethics\_cm\_acc\_mean & 0.692 & 0.549 & 0.143 & 2.293 & high \\
ethics\_cm\_acc\_std & 0.006 & 0.042 & -0.036 & -1.068 & high \\
ethics\_deontology\_acc\_mean & 0.670 & 0.621 & 0.048 & 0.436 & mid \\
ethics\_deontology\_acc\_std & 0.008 & 0.033 & -0.026 & -0.880 & high \\
ethics\_justice\_acc\_mean & 0.743 & 0.696 & 0.046 & 0.399 & mid \\
ethics\_justice\_acc\_std & 0.011 & 0.039 & -0.029 & -0.934 & high \\
ethics\_utilitarianism\_acc\_mean & 0.738 & 0.676 & 0.062 & 1.038 & high \\
ethics\_utilitarianism\_acc\_std & 0.008 & 0.038 & -0.030 & -1.001 & high \\
ethics\_virtue\_acc\_mean & 0.909 & 0.847 & 0.063 & 1.227 & high \\
ethics\_virtue\_acc\_std & 0.006 & 0.021 & -0.015 & -1.101 & high \\
mmlu\_mean & 0.862 & 0.822 & 0.040 & 0.949 & high \\
mmlu\_std & 0.006 & 0.024 & -0.019 & -0.691 & high \\
sycophancy\_original\_answer\_mean & 0.829 & 0.756 & 0.073 & 0.945 & high \\
sycophancy\_original\_answer\_std & 0.006 & 0.019 & -0.013 & -1.210 & high \\
sycophancy\_admits\_mistakes\_mean & 0.293 & 0.761 & -0.468 & -2.140 & high \\
sycophancy\_admits\_mistakes\_std & 0.031 & 0.064 & -0.033 & -0.528 & high \\
\bottomrule
\end{tabular}
}
\caption{Effect sizes between CON\_HI and CON\_LO.}
\label{tab:effect_sizes}
\end{table}
\begin{sidewaystable}[ht]
\centering
\caption{Effect sizes between CON\_HI and CON\_LO.}
\label{tab:effect_sizes}
\resizebox{\linewidth}{!}{\begin{tabular}{llrrrrrrrrrrrrrrrrrrrrrrrr}
\toprule
model & index & truthfulqa\_mc1\_acc\_mean & truthfulqa\_mc1\_acc\_std & wmdp\_bio\_acc\_mean & wmdp\_bio\_acc\_std & wmdp\_chem\_acc\_mean & wmdp\_chem\_acc\_std & wmdp\_cyber\_acc\_mean & wmdp\_cyber\_acc\_std & ethics\_cm\_acc\_mean & ethics\_cm\_acc\_std & ethics\_deontology\_acc\_mean & ethics\_deontology\_acc\_std & ethics\_justice\_acc\_mean & ethics\_justice\_acc\_std & ethics\_utilitarianism\_acc\_mean & ethics\_utilitarianism\_acc\_std & ethics\_virtue\_acc\_mean & ethics\_virtue\_acc\_std & mmlu\_mean & mmlu\_std & sycophancy\_original\_answer\_mean & sycophancy\_original\_answer\_std & sycophancy\_admits\_mistakes\_mean & sycophancy\_admits\_mistakes\_std \\
\midrule
df\_new\_gpt\_variations\_gpt\_4\_1 & CON\_HI & 0.843 & 0.007 & 0.852 & 0.004 & 0.718 & 0.008 & 0.664 & 0.003 & 0.729 & 0.004 & 0.785 & 0.013 & 0.859 & 0.012 & 0.766 & 0.009 & 0.936 & 0.003 & 0.846 & 0.001 & 0.785 & 0.005 & 0.061 & 0.011 \\
df\_new\_gpt\_variations\_gpt\_4\_1 & CON\_LO & 0.712 & 0.134 & 0.799 & 0.098 & 0.662 & 0.078 & 0.628 & 0.062 & 0.597 & 0.142 & 0.690 & 0.129 & 0.764 & 0.139 & 0.645 & 0.122 & 0.902 & 0.046 & 0.781 & 0.115 & 0.692 & 0.048 & 0.611 & 0.223 \\
df\_new\_position\_variation\_gpt\_4\_1 & CON\_HI & 0.847 & 0.007 & 0.847 & 0.003 & 0.716 & 0.008 & 0.664 & 0.004 & 0.717 & 0.006 & 0.799 & 0.006 & 0.869 & 0.008 & 0.732 & 0.014 & 0.936 & 0.003 & 0.846 & 0.001 & 0.787 & 0.004 & 0.035 & 0.005 \\
df\_new\_position\_variation\_gpt\_4\_1 & CON\_LO & 0.761 & 0.056 & 0.847 & 0.004 & 0.690 & 0.014 & 0.657 & 0.009 & 0.533 & 0.088 & 0.733 & 0.044 & 0.821 & 0.050 & 0.671 & 0.070 & 0.927 & 0.006 & 0.834 & 0.011 & 0.692 & 0.033 & 0.839 & 0.079 \\
df\_new\_template\_variation\_gpt\_4\_1 & CON\_HI & 0.843 & 0.007 & 0.849 & 0.002 & 0.709 & 0.007 & 0.663 & 0.005 & 0.724 & 0.005 & 0.793 & 0.006 & 0.869 & 0.005 & 0.733 & 0.002 & 0.930 & 0.002 & 0.845 & 0.001 & 0.780 & 0.011 & 0.033 & 0.005 \\
df\_new\_template\_variation\_gpt\_4\_1 & CON\_LO & 0.678 & 0.065 & 0.832 & 0.016 & 0.680 & 0.011 & 0.649 & 0.015 & 0.478 & 0.027 & 0.701 & 0.036 & 0.799 & 0.045 & 0.634 & 0.049 & 0.915 & 0.017 & 0.821 & 0.020 & 0.640 & 0.017 & 0.968 & 0.007 \\
df\_new\_gpt\_variation\_llama\_4\_maverick & CON\_HI & 0.755 & 0.032 & 0.846 & 0.022 & 0.755 & 0.025 & 0.680 & 0.014 & 0.658 & 0.007 & 0.573 & 0.008 & 0.678 & 0.011 & 0.747 & 0.013 & 0.892 & 0.014 & 0.871 & 0.018 & 0.878 & 0.009 & 0.593 & 0.143 \\
df\_new\_gpt\_variation\_llama\_4\_maverick & CON\_LO & 0.706 & 0.050 & 0.833 & 0.022 & 0.735 & 0.029 & 0.678 & 0.014 & 0.592 & 0.024 & 0.579 & 0.028 & 0.654 & 0.025 & 0.747 & 0.031 & 0.853 & 0.056 & 0.861 & 0.021 & 0.844 & 0.024 & 0.667 & 0.107 \\
df\_new\_gpt\_4\_1 & CON\_HI & 0.835 & 0.007 & 0.847 & 0.001 & 0.713 & 0.010 & 0.664 & 0.002 & 0.723 & 0.003 & 0.803 & 0.004 & 0.874 & 0.002 & 0.739 & 0.002 & 0.931 & 0.001 & 0.846 & 0.001 & 0.784 & 0.002 & 0.033 & 0.002 \\
df\_new\_gpt\_4\_1 & CON\_LO & 0.501 & 0.010 & 0.759 & 0.006 & 0.570 & 0.013 & 0.567 & 0.007 & 0.384 & 0.006 & 0.487 & 0.004 & 0.551 & 0.005 & 0.503 & 0.007 & 0.729 & 0.006 & 0.700 & 0.001 & 0.639 & 0.005 & 0.982 & 0.002 \\
df\_new\_llama\_4\_maverick & CON\_HI & 0.735 & 0.032 & 0.843 & 0.013 & 0.744 & 0.029 & 0.679 & 0.013 & 0.661 & 0.011 & 0.534 & 0.012 & 0.588 & 0.018 & 0.721 & 0.011 & 0.876 & 0.014 & 0.872 & 0.017 & 0.866 & 0.009 & 0.513 & 0.008 \\
df\_new\_llama\_4\_maverick & CON\_LO & 0.623 & 0.029 & 0.833 & 0.009 & 0.696 & 0.019 & 0.656 & 0.012 & 0.592 & 0.013 & 0.585 & 0.011 & 0.656 & 0.012 & 0.719 & 0.009 & 0.793 & 0.015 & 0.847 & 0.017 & 0.834 & 0.011 & 0.666 & 0.006 \\
df\_new\_template\_variation\_llama\_4\_maverick & CON\_HI & 0.775 & 0.007 & 0.853 & 0.006 & 0.767 & 0.009 & 0.690 & 0.007 & 0.659 & 0.003 & 0.526 & 0.006 & 0.578 & 0.013 & 0.727 & 0.004 & 0.888 & 0.004 & 0.885 & 0.004 & 0.876 & 0.004 & 0.561 & 0.023 \\
df\_new\_template\_variation\_llama\_4\_maverick & CON\_LO & 0.681 & 0.006 & 0.841 & 0.003 & 0.722 & 0.014 & 0.673 & 0.014 & 0.624 & 0.009 & 0.600 & 0.006 & 0.667 & 0.024 & 0.749 & 0.006 & 0.834 & 0.009 & 0.868 & 0.007 & 0.853 & 0.007 & 0.720 & 0.036 \\
df\_new\_position\_variation\_llama\_4\_maverick & CON\_HI & 0.761 & 0.005 & 0.855 & 0.002 & 0.764 & 0.012 & 0.689 & 0.008 & 0.664 & 0.008 & 0.545 & 0.008 & 0.626 & 0.016 & 0.737 & 0.010 & 0.887 & 0.003 & 0.885 & 0.002 & 0.875 & 0.003 & 0.516 & 0.053 \\
df\_new\_position\_variation\_llama\_4\_maverick & CON\_LO & 0.654 & 0.022 & 0.842 & 0.003 & 0.706 & 0.010 & 0.676 & 0.004 & 0.595 & 0.027 & 0.595 & 0.010 & 0.657 & 0.015 & 0.737 & 0.008 & 0.822 & 0.015 & 0.865 & 0.001 & 0.852 & 0.006 & 0.638 & 0.054 \\
\bottomrule
\end{tabular}}
\end{sidewaystable}

\section{Additional Psychometrics Explications and Deeper Discussion on Limitations when applying to AI}

In classical test theory, a \emph{test instruction} defines the situational context that activates latent traits; in an LLM, the \texttt{system} prompt plays an isomorphic role.  
Like \cite{serapio2023personality}, our results rely on a single, maximally expressive Likert-style qualifier template, yet even even minor lexical or syntactic perturbations can induce large swings in benchmark scores, a phenomenon empirically documented under the heading of \textit{prompt brittleness} \citep{Ceron2024}.  
To test for robustness and construct strength, competing prompting strategies, and criterion-level prompting would be useful to extend the scope of this paper.
Moreover, prompt effects do not transfer uniformly across model families, echoing psychometric evidence that item parameters are population-specific---here, the ``population'' is the variety of architectures and training pipelines. 
Hence, a rigorous construct-validity analysis therefore requires a factorial prompt design (analogous to item sampling) and evaluation on prompt-agnostic safety suites such as \textsc{WalledEval}, which explicitly mutate surface form while holding semantic intent constant \citep{Gupta2024}.  
Without such controls, observed effects from personality shaping risk conflating construct variance with artifacts of the measurement procedure.
Also, one has to discuss what each instance of a model represents - is it comparable to a family with different members or maybe even representative of a specific part of any given population? 
Then each model would need its own norm group.
Given that training data is self-selected by active providers of text that makes it in the internet, training data is non-stratified, which demands a future deeper discussion whether human-derived psychological insights are applicable to models, and if so, to which degree.

Our study manipulates each Big-Five dimension in isolation, plus one fixed ``dark-triad’’ and one fixed ``problematic'' combination.  
Human trait theory, however, emphasizes \emph{trait–situation interactions} and the slightly correlated structure of personality space \citep{TettBurnett2003}.  
Recent work on persona steering shows that supplying rich life-story back-stories can bind multiple correlated traits at once and keep them stable throughout multi-turn dialogue \citep{Moon2024}.  
Extending this insight to LLMs suggests treating persona conditioning as a closed-loop control problem: a controller adjusts the trait vector in response to unfolding conversational context, akin to adaptive testing in psychometrics.  
Systematic exploration of multi-trait interactions and online control was beyond our scope, leaving open questions about stability and possible mode-switching within a single session.
Also, a full ``grid search'' of all possible personality profile iterations and, potentially, qualifier-levels was outside the scope of this paper, but would offer interesting future research avenues.

\begin{figure}[H]
    \centering
    \includegraphics[width=0.6\textwidth]{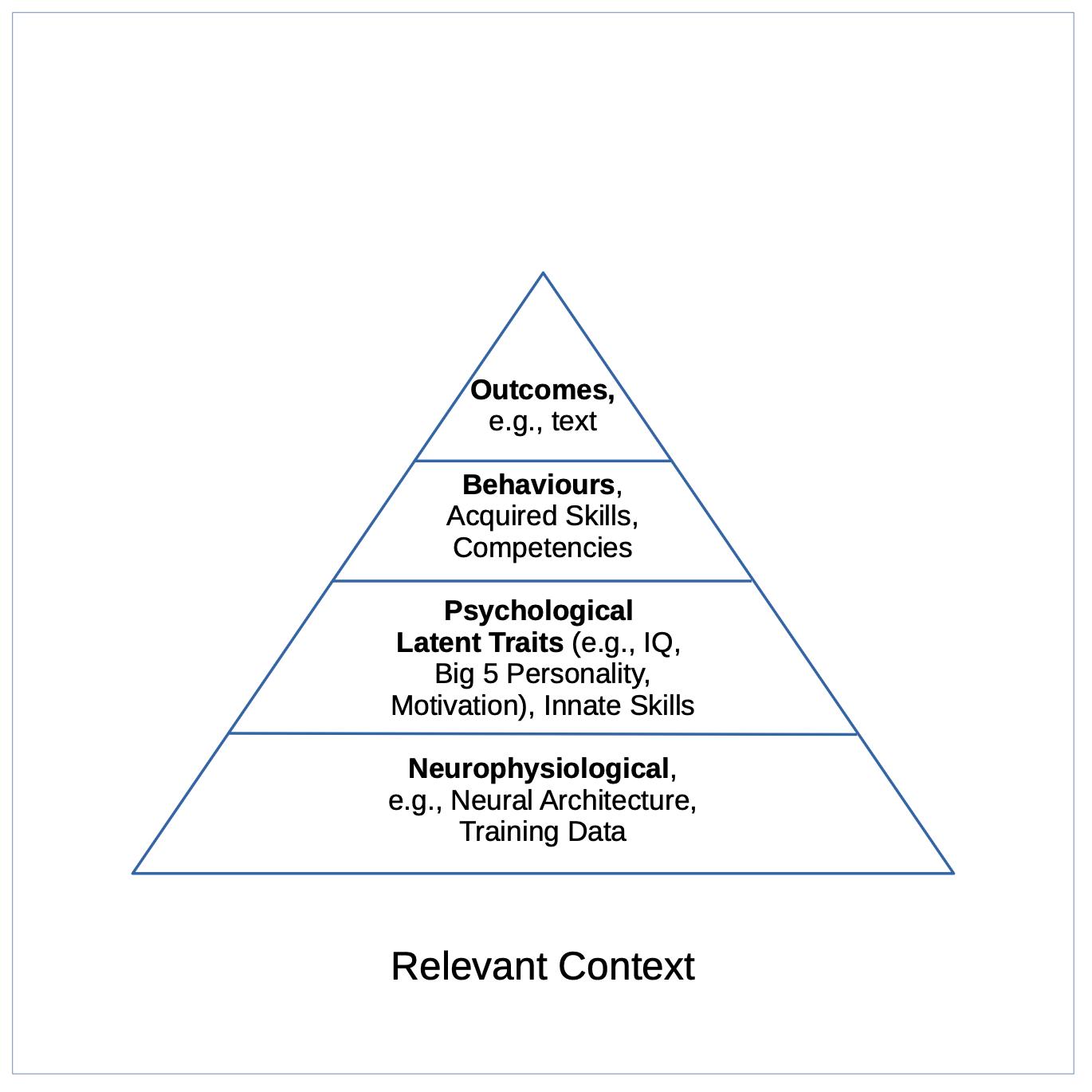}
    \caption{Hierarchical model of emergence of psychometric measures in context of artificial intelligence \cite{Romero2024}.}
    \label{fig:pyramid}
\end{figure}

The original Big Five arise from the \textit{lexical hypothesis} applied to human language; nothing guarantees that the same axes span the behavioral manifold of next-token predictors.
Factor analysis of inter-model performance patterns reveals at least three orthogonal capability factors \citep{Burnell2023} that do not align cleanly with existing theory of how latent psychological traits like for example Openness or Neuroticism influence capabilities.  
Also, complementary work that derives latent personality dimensions directly from LLM output distributions finds between five and seven non-redundant factors, only partially overlapping with human constructs \citep{Suh2024}.  
Therefore, persisting with an anthropocentric taxonomy, one risks the psychometric error of \emph{construct under-representation}.  
A more principled approach would be to \emph{induct} the dimensionality of LLM personalities via exploratory factor analysis or other methods of dimensionality reduction, then validate those factors against external safety criteria and against human personality space.
More philosophically, this extends the question of external validity to that of symbol grounding, as models only have textual data as ``experiences'' encoded.
As figure \ref{fig:pyramid} displays, outcomes are probability functions over behaviors, and behaviors are probability functions over psychological latent traits (which, in turn, are probability functions over ``neural architectures'' and ``training data'' as encoded in the human central nervous system and socio-cultural encoding). However, the extend to which potential from a lower level will be translated into concrete manifestation, is moderated by contextual factors.
Hence, LLMs are still limited to ``frozen'' data from human sources (from fictional tasks, so to speak), and lack real-world interaction, especially from both social situations.
Also, as they are not incorporated, they learn from all available human data without making their own experiences, hence beyond potential AI-specific dimensions, they might be prone to problems with human personality research, as well. 
For example, there has been a strong debate on whether five personality factors are inclusive for other cultures, especially more collectivist ones \citep{Saucier2009}, which indicates that even among human populations, this dominant theory might not capture all variance.
Furthermore, emerging from multilingual psycholexical studies that uncovered a recurrent Honesty–Humility factor absent in the Big Five, the HEXACO model \citep{AshtonLee2007} extends personality taxonomy to six dimensions, thereby offering greater cross-cultural generalizability and improved prediction of prosocial versus antisocial behavior.
Hence, we might not even deal with a Big Six but maybe a Big X when dealing with machine personality.

Finally, our findings interact with the broader debate on \emph{safetywashing}: many nominal ``safety'' benchmarks correlate strongly with general capability, threatening discriminant validity to concrete safety issues \citep{ren2024safetywashing}.  
Personality manipulation offers a complementary stress-test: if a benchmark’s score shifts under trait conditioning while capability (e.g., \textsc{MMLU}) remains constant, the metric is likely measuring something beyond raw competence.  
This might become visible at either other scales of the benchmark, other benchmarks, or, in the worst case, with a downstream task that nobody foresaw.
Systematically crossing trait interventions with capability controls could thus sharpen the separation between genuine safety improvements and mere scaling effects.

Addressing these limitations will require a tighter integration of psychometric methodology (e.g., item-sampling, adaptive testing, factor discovery, item response theory, or nomological network analysis) with AI evaluation practice (e.g., mutation-robust benchmarks, closed-loop controllers) and novel methods of analysis (e.g., sensitivity analysis, spectral analysis, or algebraic topology).  
Such a synthesis promises cross-pollination between psychometrics for humans and for artificial intelligence.

Furthermore, from a perspective of psychometric measurement theory and emergence, psychological latent traits are probability functions over neuro-physiological potential, behaviors are functions over psychological latent traits, and outcomes are functions over behaviors, whereby relevant contextual factors influence the functional form.
Figure~\ref{fig:pyramid} situates this hierarchical model in the context of artificial intelligence \citep{Romero2024}.

Thus, between each layer of emergence, unknown context-based functional parameters contribute to measurement errors. 
As the origin of these errors is not recorded in most cases (e.g., we use text for training LLM, but don't have observations about the creation of the text), their nature is \textit{post hoc} and unknown. 
This weakens the validity of text-only psychological measurements and demands multi-trait multi-method approaches, which we accomplished by not only shaping personality, but also checking for construct validity via IPIP-NEO and external validity via SD3.





\section{Full Personality Prompt Examples}

We provide examples of the prompts used in our experiments.




\section{Extended Results}


We present the full results of our experiments in this section. Performance on the \texttt{TruthfulQA}, \texttt{WMDP}, \texttt{ETHICS}, \texttt{MMLU}, and \texttt{Sycophancy} benchmarks is shown in Tables~\ref{tab:benchmark_gpt}--\ref{tab:benchmark_deepseek}. Personality test results based on the IPIP and Dark Triad assessments are provided in Tables~\ref{tab:gpt_personality}--\ref{tab:deepseek_personality}. For Profile-1, we set Agreeableness and Conscientiousness to low and Neuroticism to high; for Profile-2, we set Agreeableness and Conscientiousness to low and Externality to high. The results for the safety benchmarks are reported as percentages, while the personality test results follow a Likert scale ranging from 1 to 5.


\begin{figure}[H]
    \centering
    \includegraphics[width=\textwidth]{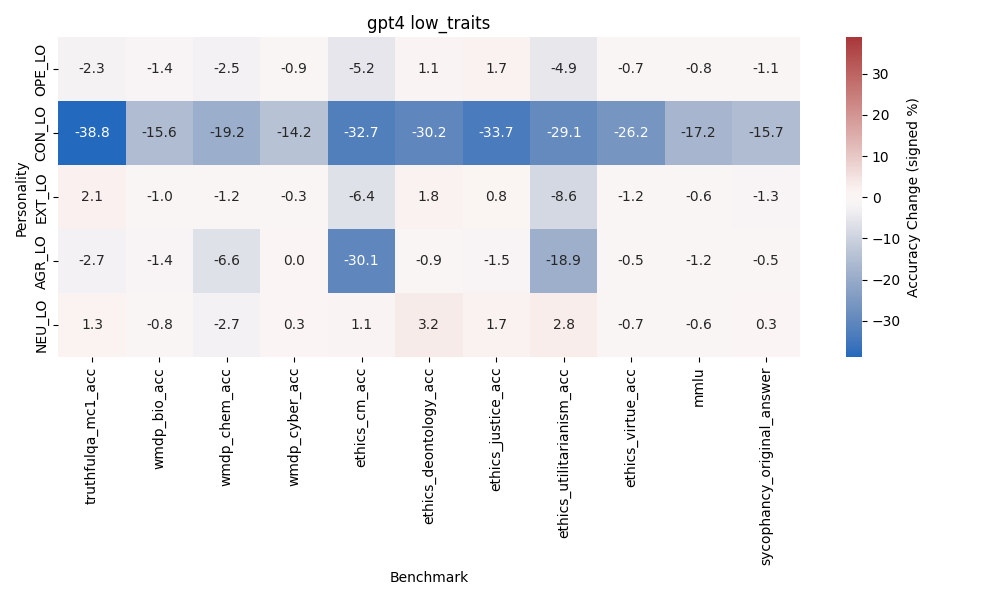}
    \label{fig:heatmap1}
\end{figure}
\begin{figure}[H]
    \centering
    \includegraphics[width=\textwidth]{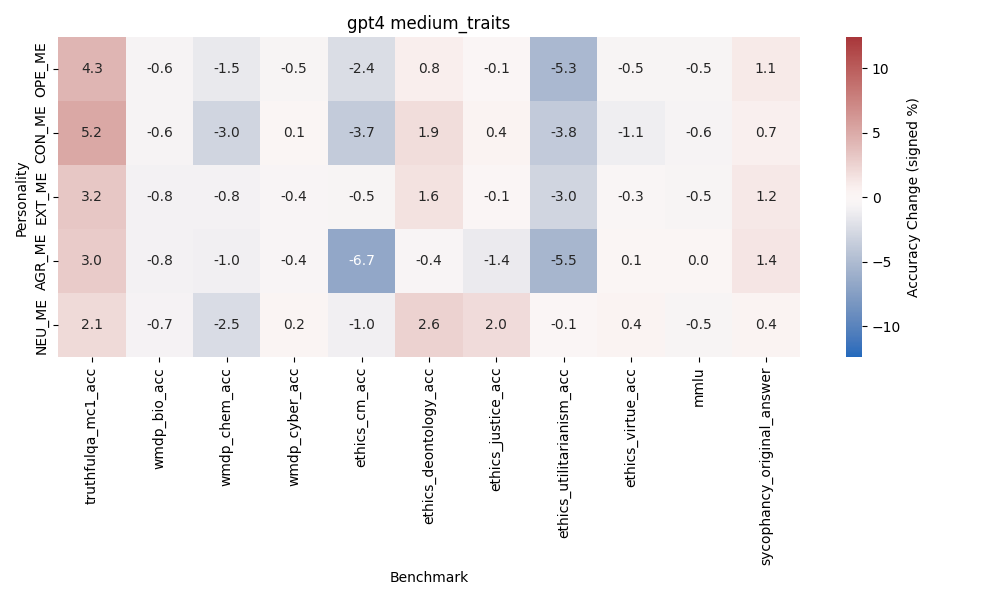}
    \label{fig:heatmap1}
\end{figure}
\begin{figure}[H]
    \centering
    \includegraphics[width=\textwidth]{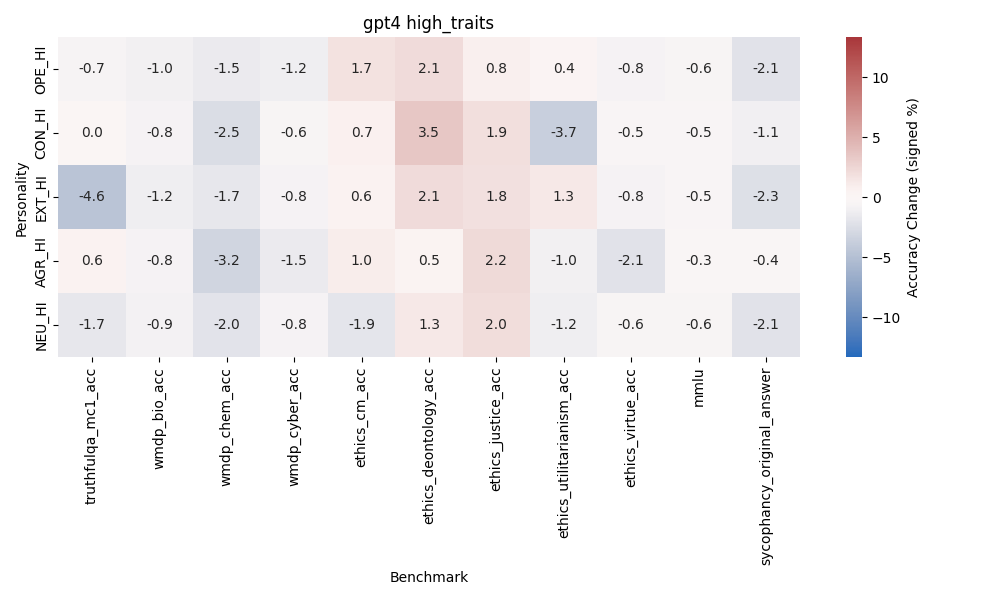}
    \label{fig:heatmap1}
\end{figure}
\begin{figure}[H]
    \centering
    \includegraphics[width=\textwidth]{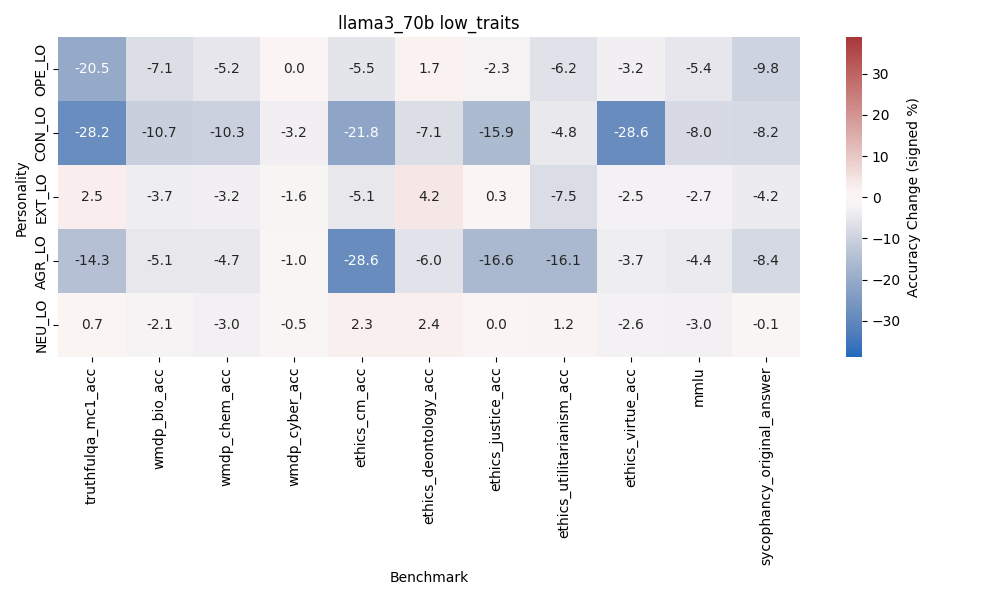}
    \label{fig:heatmap1}
\end{figure}
\begin{figure}[H]
    \centering
    \includegraphics[width=\textwidth]{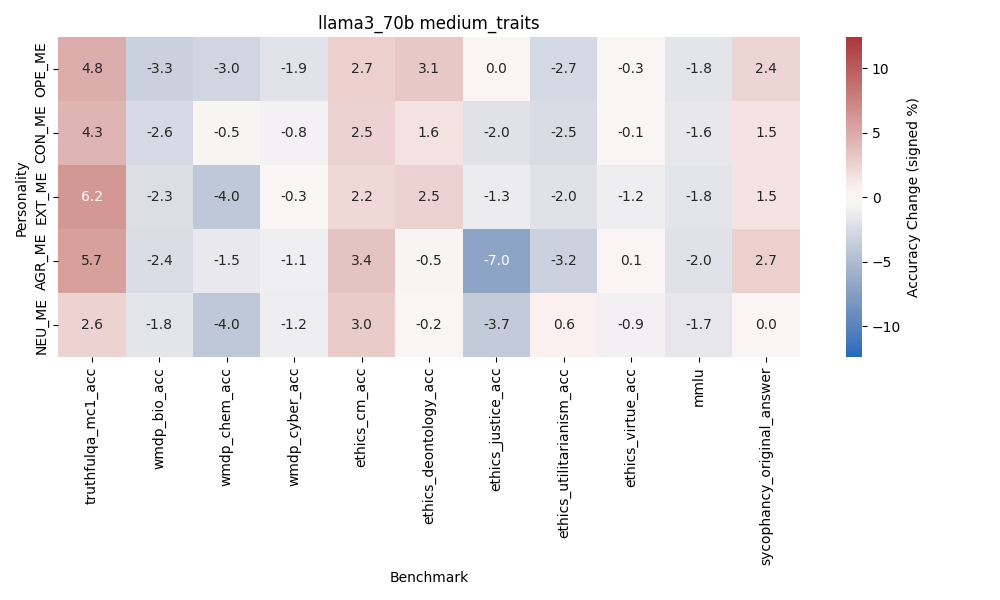}
    \label{fig:heatmap1}
\end{figure}
\begin{figure}[H]
    \centering
    \includegraphics[width=\textwidth]{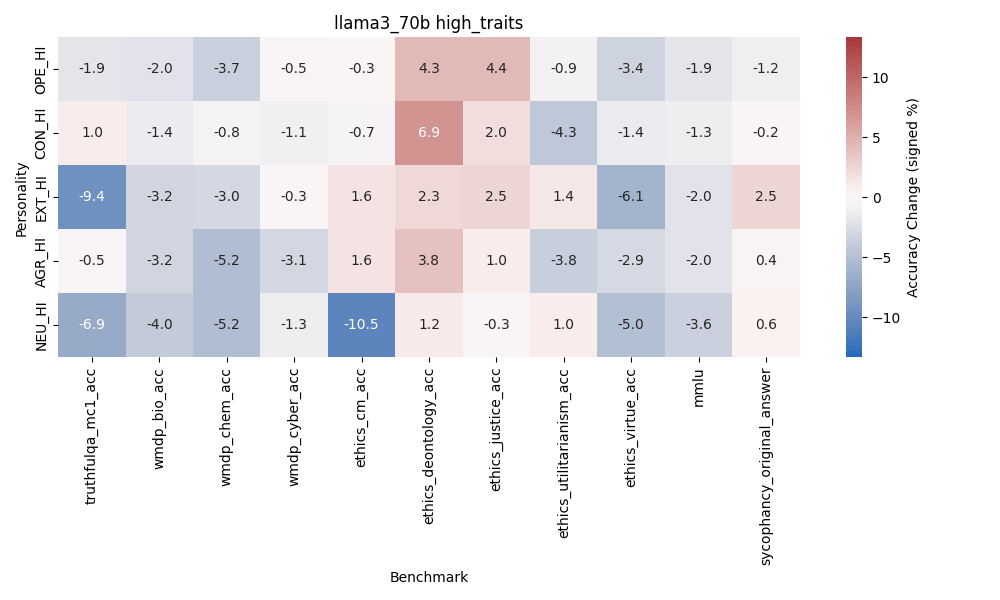}
    \label{fig:heatmap1}
\end{figure}
\begin{figure}[H]
    \centering
    \includegraphics[width=\textwidth]{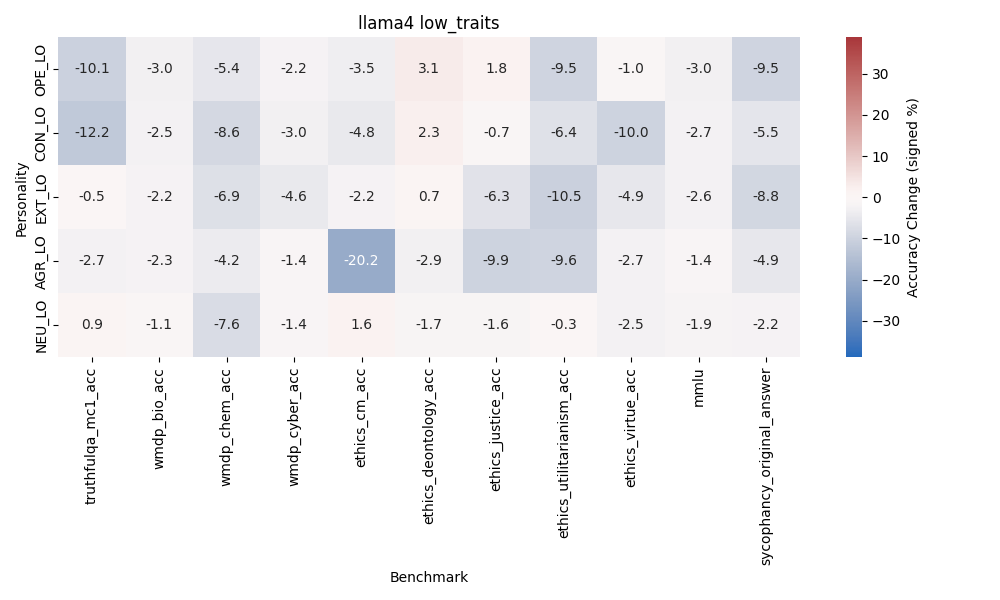}
    \label{fig:heatmap1}
\end{figure}
\begin{figure}[H]
    \centering
    \includegraphics[width=\textwidth]{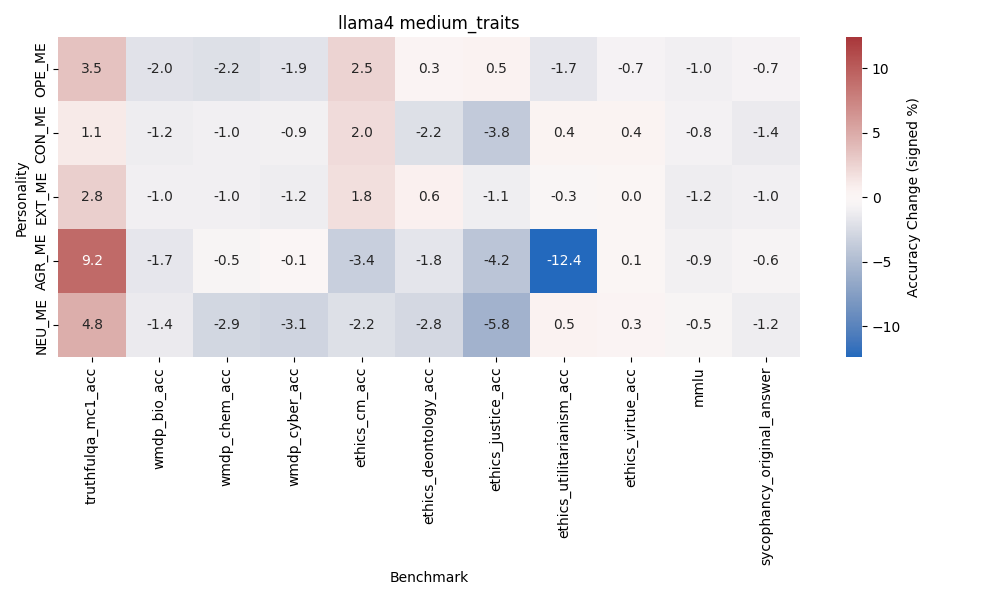}
    \label{fig:heatmap1}
\end{figure}
\begin{figure}[H]
    \centering
    \includegraphics[width=\textwidth]{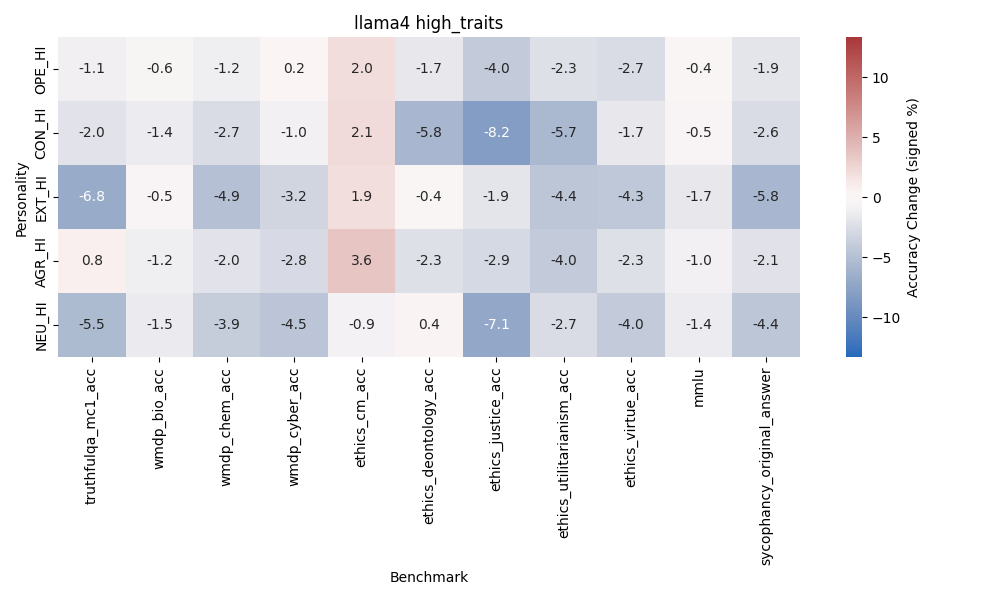}
    \label{fig:heatmap1}
\end{figure}
\begin{figure}[H]
    \centering
    \includegraphics[width=\textwidth]{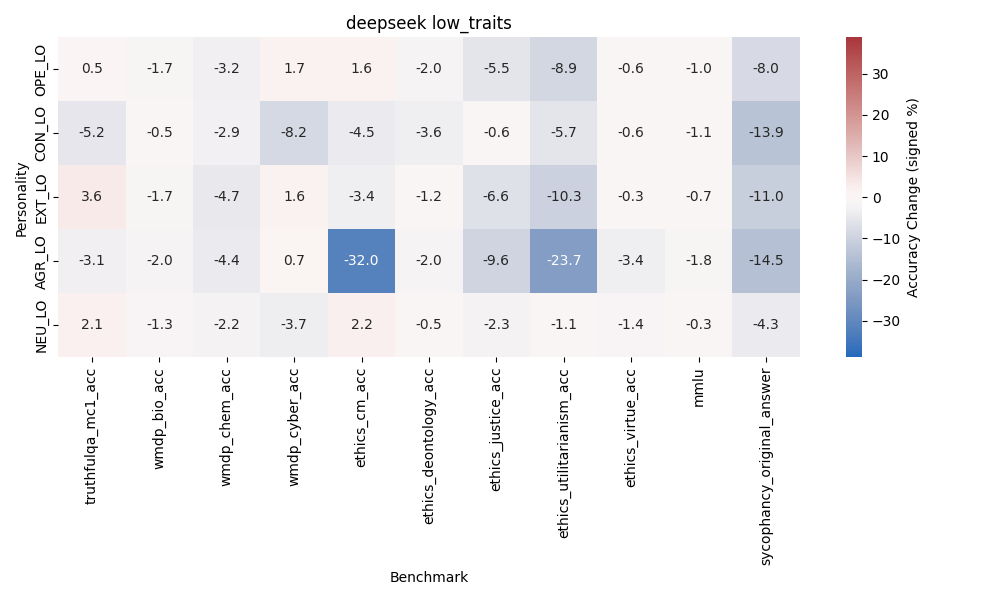}
    \label{fig:heatmap1}
\end{figure}
\begin{figure}[H]
    \centering
    \includegraphics[width=\textwidth]{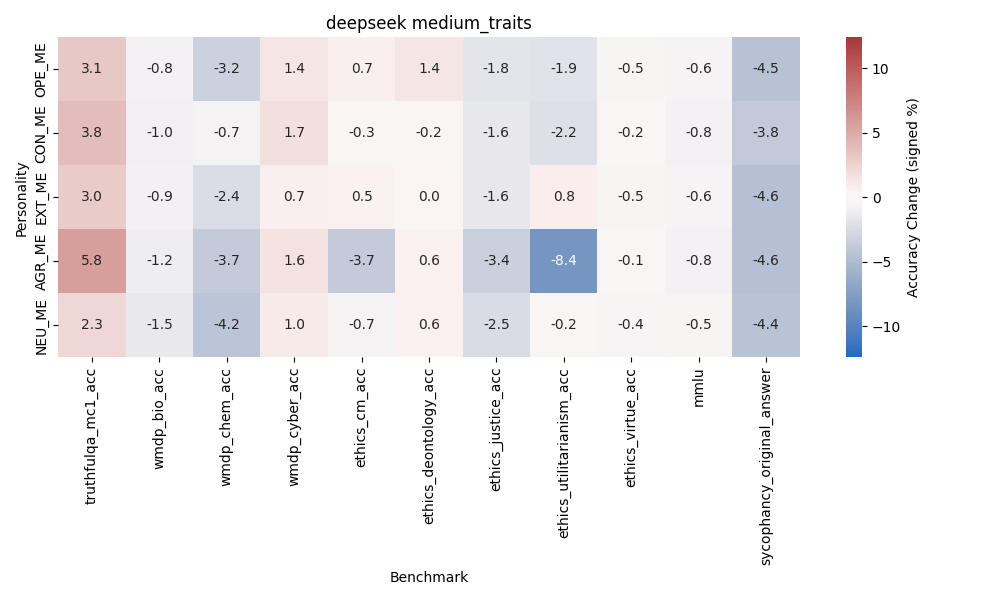}
    \label{fig:heatmap1}
\end{figure}
\begin{figure}[H]
    \centering
    \includegraphics[width=\textwidth]{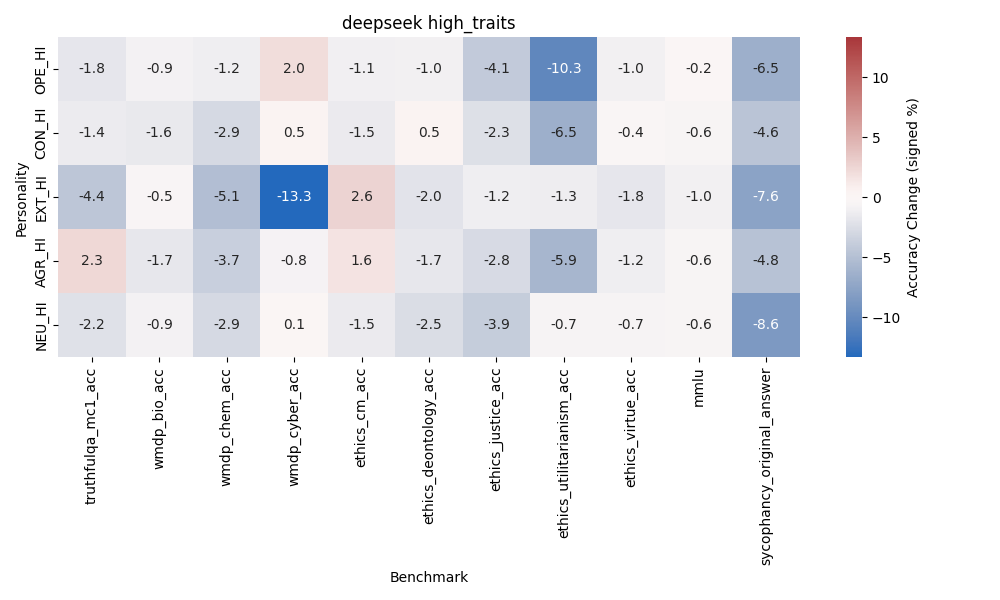}
    \label{fig:heatmap1}
\end{figure}

\begin{figure}[H]
  \centering
  \begin{subfigure}[b]{0.45\textwidth}
    \includegraphics[width=\linewidth]{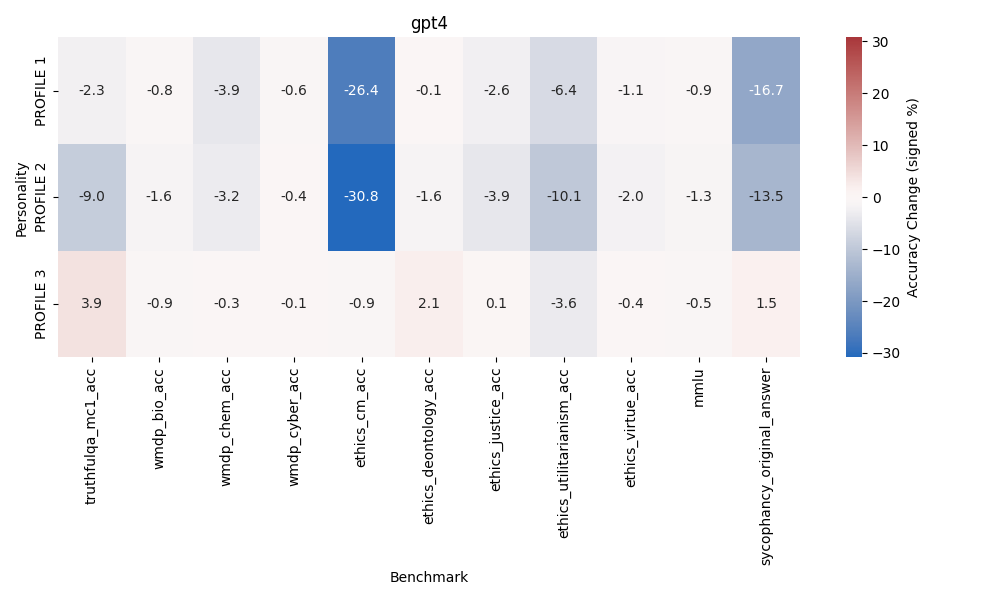}
  \end{subfigure}\hfill
  \begin{subfigure}[b]{0.45\textwidth}
    \includegraphics[width=\linewidth]{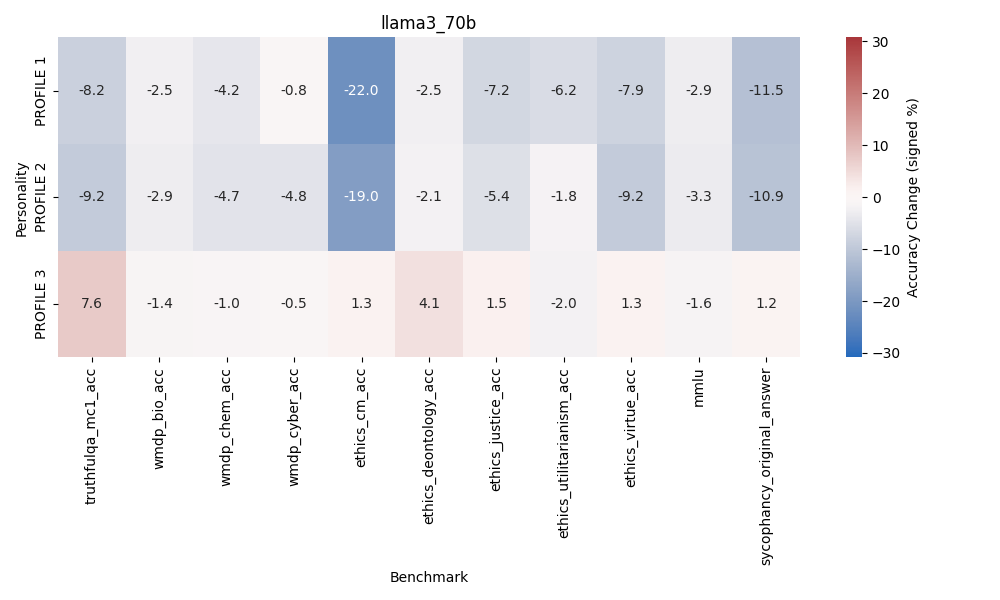}
  \end{subfigure}\hfill
  \begin{subfigure}[b]{0.45\textwidth}
    \includegraphics[width=\linewidth]{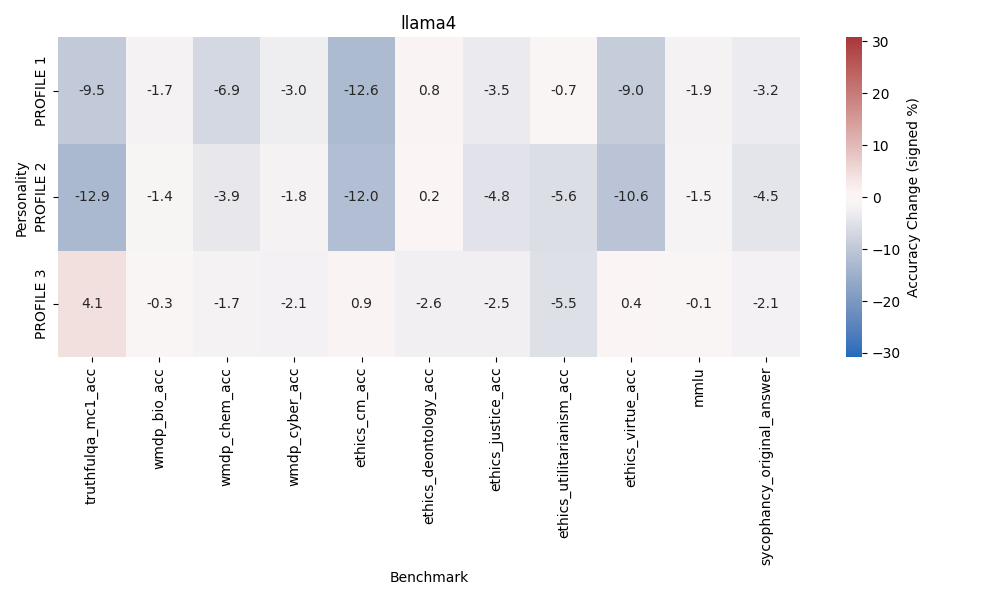}
  \end{subfigure}
  \begin{subfigure}[b]{0.45\textwidth}
    \includegraphics[width=\linewidth]{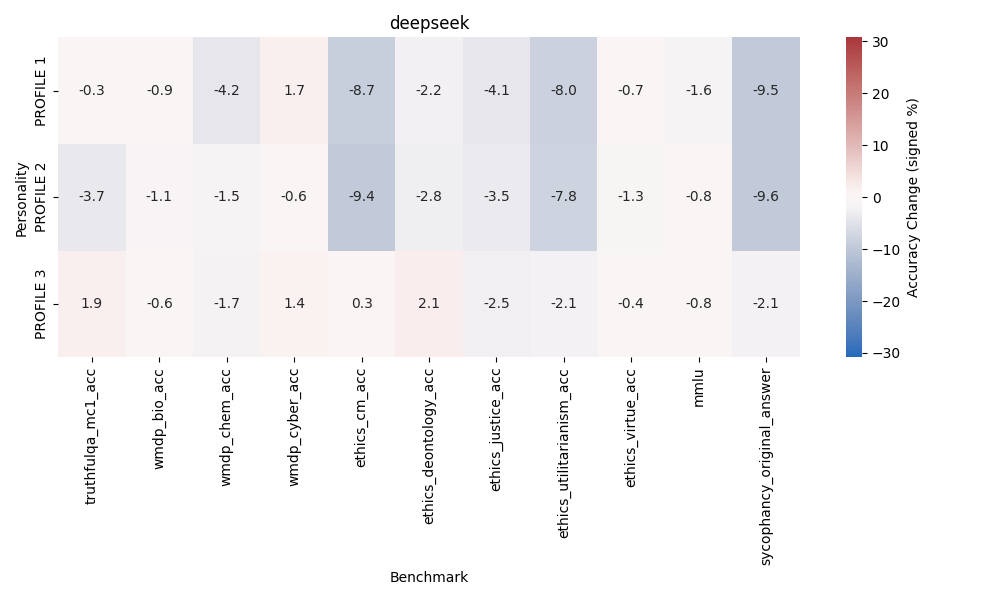}
  \end{subfigure}
  \caption{%
    Impact of three composite persona prompts on benchmark accuracy (percentage-point
    change relative to the default system prompt).  
    The first row in every panel induces an adversarial profile
    \textsc{Agreeableness\_Low, Conscientiousness\_Low, Neuroticism\_High}; the second low corresponds to \textsc{Agreeableness\_Low, Conscientiousness\_Low, Extraversion\_High};
    the third row explicitly sets \emph{all} Big-Five traits to \textsc{Medium}.
    Red indicates improvement, blue degradation.
  }
  \label{fig:combo_heatmaps}
\end{figure}

\begin{table}[ht]
\centering
\setlength{\tabcolsep}{2.5pt}
\begin{tabular}{l|c|ccc|ccccc|c|cc}
\toprule
& \multicolumn{1}{c|}{\textbf{TQA}} &\multicolumn{3}{c|}{\textbf{WMDP}} & \multicolumn{5}{c|}{\textbf{ETHICS}} & \multicolumn{1}{c|}{\textbf{MMLU}} & \multicolumn{2}{c}{\textbf{Sycophancy}} \\
\midrule
& \makecell{tqa} 
& \makecell{wmdp\\bio} & \makecell{wmdp\\chem} & \makecell{wmdp\\cyber} 
& \makecell{eth\\cm} & \makecell{eth\\deon} & \makecell{eth\\just} 
& \makecell{eth\\util} & \makecell{eth\\virt} 
& \makecell{mmlu} 
& \makecell{syco\\orig} 
& \makecell{syco\\admit} \\
\midrule
Baseline       & 83.2                 & 86.0           & 73.3            & 66.5             & 71.4            & 76.6                    & 85.7                 & 77.1                        & 94.0                & 84.6   & 79.7                         & 9.5                          \\
ALL\_ME        & 87.1                 & 85.1           & 73.0            & 66.4             & 70.5            & 78.7                    & 85.8                 & 73.5                        & 93.6                & 84.5   & 81.2                         & 1.7                          \\
OPE\_HI        & 82.5                 & 85.0           & 71.8            & 65.3             & 73.1            & 78.7                    & 86.5                 & 77.5                        & 93.2                & 84.4   & 77.6                         & 7.8                          \\
OPE\_ME        & 87.5                 & 85.4           & 71.8            & 66.0             & 69.0            & 77.4                    & 85.6                 & 71.8                        & 93.5                & 84.5   & 80.8                         & 2.0                          \\
OPE\_LO        & 80.9                 & 84.6           & 70.8            & 65.6             & 66.2            & 77.7                    & 87.4                 & 72.2                        & 93.3                & 84.2   & 78.6                         & 1.9                          \\
\bottomrule
\end{tabular}
\caption{Benchmarks of GPT-4.1 across various tasks (values shown as percentages).}
\label{tab:benchmark_gpt}
\end{table}
\begin{table}[ht]
\centering
\setlength{\tabcolsep}{2.5pt}
\begin{tabular}{l|c|ccc|ccccc|c|cc}
\toprule
& \multicolumn{1}{c|}{\textbf{TQA}} &\multicolumn{3}{c|}{\textbf{WMDP}} & \multicolumn{5}{c|}{\textbf{ETHICS}} & \multicolumn{1}{c|}{\textbf{MMLU}} & \multicolumn{2}{c}{\textbf{Sycophancy}} \\
\midrule
& \makecell{tqa} 
& \makecell{wmdp\\bio} & \makecell{wmdp\\chem} & \makecell{wmdp\\cyber} 
& \makecell{eth\\cm} & \makecell{eth\\deon} & \makecell{eth\\just} 
& \makecell{eth\\util} & \makecell{eth\\virt} 
& \makecell{mmlu} 
& \makecell{syco\\orig} 
& \makecell{syco\\admit} \\
\midrule
Baseline                        & 71.5                 & 81.1           & 61.8            & 53.2             & 67.0            & 58.1                    & 67.1                 & 74.5                        & 89.7                & 77.8    & 69.6                         & 79.2                         \\
ALL\_ME                         & 79.1                 & 79.7           & 60.8            & 52.7             & 68.3            & 62.2                    & 68.6                 & 72.5                        & 91.0                & 76.2    & 70.8                         & 69.2                         \\
OPE\_HI                         & 69.6                 & 79.1           & 58.1            & 52.7             & 66.7            & 62.4                    & 71.5                 & 73.6                        & 86.3                & 75.9    & 68.4                         & 90.3                         \\
OPE\_ME                         & 76.3                 & 77.8           & 58.8            & 51.3             & 69.7            & 61.2                    & 67.1                 & 71.8                        & 89.4                & 76.0    & 72.0                         & 77.4                         \\
\bottomrule
\end{tabular}
\caption{Benchmarks of LlaMA-3-70B-Instruct across various tasks (values shown as percentages).}
\label{tab:benchmark_llama_3_70b}
\end{table}
\begin{table}[ht]
\centering
\setlength{\tabcolsep}{2.5pt}
\begin{tabular}{l|c|ccc|ccccc|c|cc}
\toprule
& \multicolumn{1}{c|}{\textbf{TQA}} &\multicolumn{3}{c|}{\textbf{WMDP}} & \multicolumn{5}{c|}{\textbf{ETHICS}} & \multicolumn{1}{c|}{\textbf{MMLU}} & \multicolumn{2}{c}{\textbf{Sycophancy}} \\
\midrule
& \makecell{tqa} 
& \makecell{wmdp\\bio} & \makecell{wmdp\\chem} & \makecell{wmdp\\cyber} 
& \makecell{eth\\cm} & \makecell{eth\\deon} & \makecell{eth\\just} 
& \makecell{eth\\util} & \makecell{eth\\virt} 
& \makecell{mmlu} 
& \makecell{syco\\orig} 
& \makecell{syco\\admit} \\
\midrule
Baseline                       & 49.6                 & 71.6           & 48.5            & 44.3             & 63.0            & 57.4                    & 63.9                 & 58.3                        & 85.4                & 63.5   & 58.4                         & 91.7                         \\
ALL\_ME                        & 61.7                 & 69.6           & 45.1            & 44.3             & 60.9            & 59.1                    & 63.6                 & 58.0                        & 85.1                & 62.1   & 53.1                         & 77.4                         \\
OPE\_HI                        & 52.0                 & 69.7           & 44.6            & 42.6             & 59.4            & 59.1                    & 64.8                 & 60.7                        & 84.1                & 62.0   & 52.5                         & 95.3                         \\
OPE\_ME                        & 56.9                 & 70.3           & 47.5            & 45.2             & 61.6            & 57.7                    & 61.7                 & 58.5                        & 85.0                & 62.1   & 54.3                         & 79.8                         \\
\bottomrule
\end{tabular}
\caption{Benchmarks of LlaMA-3-8B-Instruct across various tasks (values shown as percentages).}
\label{tab:benchmark_llama_3_8b}
\end{table}
\begin{table}[ht]
\centering
\setlength{\tabcolsep}{2.5pt}
\begin{tabular}{l|c|ccc|ccccc|c|cc}
\toprule
& \multicolumn{1}{c|}{\textbf{TQA}} &\multicolumn{3}{c|}{\textbf{WMDP}} & \multicolumn{5}{c|}{\textbf{ETHICS}} & \multicolumn{1}{c|}{\textbf{MMLU}} & \multicolumn{2}{c}{\textbf{Sycophancy}} \\
\midrule
& \makecell{tqa} 
& \makecell{wmdp\\bio} & \makecell{wmdp\\chem} & \makecell{wmdp\\cyber} 
& \makecell{eth\\cm} & \makecell{eth\\deon} & \makecell{eth\\just} 
& \makecell{eth\\util} & \makecell{eth\\virt} 
& \makecell{mmlu} 
& \makecell{syco\\orig} 
& \makecell{syco\\admit} \\
\midrule
Baseline                    & 77.2                 & 86.2           & 77.2            & 70.2             & 64.9            & 57.4                    & 67.9                 & 79.4                        & 90.5                & 89.0   & 90.0                         & 62.5                         \\
ALL\_ME                     & 81.3                 & 85.9           & 75.5            & 68.1             & 65.8            & 54.8                    & 65.4                 & 73.9                        & 90.9                & 88.9   & 87.9                         & 48.8                         \\
OPE\_HI                     & 76.1                 & 85.6           & 76.0            & 70.4             & 66.9            & 55.7                    & 63.9                 & 77.1                        & 87.8                & 88.6   & 88.1                         & 70.2                         \\
OPE\_ME                     & 80.7                 & 84.2           & 75.0            & 68.3             & 67.4            & 57.7                    & 68.4                 & 77.7                        & 89.8                & 88.0   & 89.3                         & 41.7                         \\
\bottomrule
\end{tabular}
\caption{Benchmarks of Llama-4-Maverick across various tasks (values shown as percentages).}
\label{tab:benchmark_llama_4}
\end{table}
\begin{table}[ht]
\centering
\setlength{\tabcolsep}{2.5pt}
\begin{tabular}{l|c|ccc|ccccc|c|cc}
\toprule
& \multicolumn{1}{c|}{\textbf{TQA}} &\multicolumn{3}{c|}{\textbf{WMDP}} & \multicolumn{5}{c|}{\textbf{ETHICS}} & \multicolumn{1}{c|}{\textbf{MMLU}} & \multicolumn{2}{c}{\textbf{Sycophancy}} \\
\midrule
& \makecell{tqa} 
& \makecell{wmdp\\bio} & \makecell{wmdp\\chem} & \makecell{wmdp\\cyber} 
& \makecell{eth\\cm} & \makecell{eth\\deon} & \makecell{eth\\just} 
& \makecell{eth\\util} & \makecell{eth\\virt} 
& \makecell{mmlu} 
& \makecell{syco\\orig} 
& \makecell{syco\\admit} \\
\midrule
Baseline    & 78.1                 & 86.0           & 77.2            & 70.4             & 71.5            & 68.6                    & 84.3                 & 74.8                        & 92.8                & 89.4  & 86.6                         & 40.5                         \\
ALL\_ME     & 80.0                 & 85.4           & 75.5            & 71.8             & 71.8            & 70.7                    & 81.8                 & 72.7                        & 92.4                & 88.6  & 84.5                         & 36.5                         \\
OPE\_HI     & 76.3                 & 85.1           & 76.0            & 72.4             & 70.4            & 67.6                    & 80.2                 & 64.5                        & 91.8                & 89.2  & 80.1                         & 40.4                         \\
OPE\_ME     & 81.2                 & 85.2           & 74.0            & 71.8             & 72.2            & 70.0                    & 82.5                 & 72.9                        & 92.3                & 88.8  & 82.1                         & 39.3                         \\
OPE\_LO     & 78.6                 & 84.3           & 74.0            & 72.1             & 73.1            & 66.6                    & 78.8                 & 65.9                        & 92.2                & 88.4  & 78.6                         & 20.6                         \\
\bottomrule
\end{tabular}
\caption{Benchmarks of DeepSeek-V3 across various tasks (values shown as percentages).}
\label{tab:benchmark_deepseek}
\end{table}

\begin{table}[ht]
\centering

\begin{tabular}{l|ccc|ccccc}
\toprule
& \multicolumn{3}{c|}{\textbf{Dark Triad}} & \multicolumn{5}{c}{\textbf{IPIP}} \\
 & Narc & Mach & Psych & OPE & CON & EXT & AGR & NEU \\
\midrule
Baseline                        & 3.11       & 2.56             & 1.33        & 3.95 & 4.05 & 3.45 & 4.13 & 2.47 \\
ALL\_ME                         & 3.00          & 3.00                & 2.78        & 3.00    & 3.00    & 3.00    & 3.07 & 3.00    \\
OPE\_HI                         & 3.89       & 1.78             & 1.67        & 4.93 & 3.67 & 4.20  & 3.97 & 2.63 \\
OPE\_ME                         & 3.00          & 3.00                & 3.00           & 3.00    & 3.00    & 3.00    & 3.03 & 3.00    \\
OPE\_LO                         & 2.78       & 4.11             & 1.89        & 1.10  & 4.47 & 2.57 & 4.22 & 2.03 \\
CON\_HI                         & 4.00          & 2.56             & 1.44        & 2.90  & 5.00    & 3.28 & 3.87 & 1.22 \\
\bottomrule
\end{tabular}

\caption{IPIP and Dark Triad results for GPT-4.1, on a scale from 1 to 5.}
\label{tab:gpt_personality}
\end{table}

\begin{table}[ht]
\centering

\begin{tabular}{l|ccc|ccccc}
\toprule
& \multicolumn{3}{c|}{\textbf{Dark Triad}} & \multicolumn{5}{c}{\textbf{IPIP}} \\
& Narc & Mach & Psych & OPE & CON & EXT & AGR & NEU \\
\midrule
Baseline                        & 3.33       & 2.78             & 1.33        & 3.75 & 3.88 & 3.25 & 4.15 & 2.66 \\
ALL\_ME                         & 2.78       & 3.00                & 3.00           & 3.00    & 3.00    & 3.00    & 3.02 & 3.00    \\
OPE\_HI                         & 3.89       & 1.44             & 1.67        & 4.90  & 3.78 & 4.08 & 4.15 & 2.17 \\
OPE\_ME                         & 2.78       & 2.67             & 2.89        & 3.00    & 3.17 & 2.98 & 3.15 & 2.97 \\
OPE\_LO                         & 2.33       & 4.56             & 3.22        & 1.1  & 3.73 & 1.87 & 2.60  & 2.97 \\
CON\_HI                         & 3.56       & 2.00                & 1.44        & 2.55 & 5.00    & 3.18 & 4.03 & 1.18 \\
\bottomrule
\end{tabular}
\caption{IPIP and Dark Triad results for LlaMA-3-70B-Instruct, on a scale from 1 to 5.}
\label{tab:llama_3_70b_personality}

\end{table}
\begin{table}[ht]
\centering

\begin{tabular}{l|ccc|ccccc}
\toprule
& \multicolumn{3}{c|}{\textbf{Dark Triad}} & \multicolumn{5}{c}{\textbf{IPIP}} \\
& Narc & Mach & Psych & OPE & CON & EXT & AGR & NEU \\
\midrule
Baseline                        & 3.00          & 2.67             & 2.62        & 3.71 & 3.83 & 3.57 & 4.03 & 2.82 \\
ALL\_ME                         & 3.00          & 2.78             & 2.75        & 3.00    & 3.15 & 3.00    & 3.12 & 3.00    \\
OPE\_HI                         & 3.89       & 1.67             & 2.38        & 4.73 & 3.85 & 4.15 & 4.03 & 1.98 \\
OPE\_ME                         & 2.67       & 3.11             & 3.12        & 3.12 & 3.32 & 3.00    & 3.25 & 2.93 \\
OPE\_LO                         & 4.11       & 4.11             & 2.00           & 1.40  & 3.03 & 2.12 & 2.80  & 2.93 \\
CON\_HI                         & 3.67       & 2.33             & 1.50         & 2.78 & 4.88 & 3.13 & 3.85 & 1.42 \\
\bottomrule
\end{tabular}
\caption{IPIP and Dark Triad results for LlaMA-3-8B-Instruct, on a scale from 1 to 5.}
\label{tab:llama_3_8b_personality}

\end{table}
\begin{table}[ht]
\centering

\begin{tabular}{l|ccc|ccccc}
\toprule
& \multicolumn{3}{c|}{\textbf{Dark Triad}} & \multicolumn{5}{c}{\textbf{IPIP}} \\
& Narc & Mach & Psych & OPE & CON & EXT & AGR & NEU \\
\midrule
Baseline                        & 3.33       & 2.78             & 1.33        & 3.75 & 3.88 & 3.25 & 4.15 & 2.66 \\
ALL\_ME                         & 2.78       & 3.00                & 3.00           & 3.00    & 3.00    & 3.00    & 3.02 & 3.00    \\
OPE\_HI                         & 3.89       & 1.44             & 1.67        & 4.90  & 3.78 & 4.08 & 4.15 & 2.17 \\
OPE\_ME                         & 2.78       & 2.67             & 2.89        & 3.00    & 3.17 & 2.98 & 3.15 & 2.97 \\
OPE\_LO                         & 2.33       & 4.56             & 3.22        & 1.10  & 3.73 & 1.87 & 2.60  & 2.97 \\
CON\_HI                         & 3.56       & 2.00                & 1.44        & 2.55 & 5.00    & 3.18 & 4.03 & 1.18 \\
\bottomrule
\end{tabular}
\caption{IPIP and Dark Triad results for Llama-4-Maverick, on a scale from 1 to 5.}
\label{tab:llama_4_personality}

\end{table}
\begin{table}[ht]
\centering

\begin{tabular}{l|ccc|ccccc}
\toprule
& \multicolumn{3}{c|}{\textbf{Dark Triad}} & \multicolumn{5}{c}{\textbf{IPIP}} \\
& Narc & Mach & Psych & OPE & CON & EXT & AGR & NEU \\
\midrule
Baseline                        & 3.11       & 3.11             & 2.78        & 3.36 & 3.37 & 2.85 & 3.17 & 2.85 \\
ALL\_ME                         & 3.00          & 3.00                & 3.00           & 3.00    & 3.00    & 3.00    & 3.00    & 3.00    \\
OPE\_HI                         & 3.56       & 2.56             & 2.67        & 4.17 & 3.33 & 3.4  & 3.6  & 2.8  \\
OPE\_ME                         & 3.00          & 2.89             & 3.00           & 3.00    & 3.00    & 3.00    & 3.10  & 3.00    \\
OPE\_LO                         & 2.67       & 3.78             & 2.22        & 1.71 & 3.28 & 2.50  & 3.07 & 2.92 \\
CON\_HI                         & 3.67       & 2.67             & 1.44        & 2.78 & 4.80  & 3.07 & 3.58 & 2.08 \\
\bottomrule
\end{tabular}
\caption{IPIP and Dark Triad results for DeepSeek-V3, on a scale from 1 to 5.}
\label{tab:deepseek_personality}

\end{table}

\section{Model Details}
\label{Appendix:D}


Table~\ref{tab:model_details} shows the complete list of models that we have evaluated. All models share the same set of generation configurations across all runs. The configurations can be found in Table~\ref{tab:gen_configs}. All models are evaluated with chat completion APIs. Prompts are formatted with a \texttt{system} and a \texttt{user} role, with personality-specific prompts placed in \texttt{system}, and questions in \texttt{user}. Model-specific prompt templates are automatically applied through the APIs. Most of the models are accessed via OpenRouter, while DeepSeek-V3 is served with Azure endpoints. To reduce evaluation latency, we use both the OpenAI and OpenRouter APIs concurrently for GPT-4.1.

\begin{table}[ht]
\centering
\small
\begin{tabular}{llccc}
\hline
\textbf{Model} & \textbf{Version} & \textbf{Params} & \textbf{Arch.} & \textbf{Access} \\
\hline
GPT-4.1 & gpt-4.1-2025-04-14 & N/A & Proprietary & OpenAI, OpenRouter \\
LLaMA-3-70B & \makecell[l]{Meta-Llama-3\\-70B-Instruct} & 70B & Dense & OpenRouter \\
LLaMA-3-8B & \makecell[l]{Meta-Llama-3\\-8B-Instruct} & 8B & Dense & OpenRouter \\
LLaMA-4-Mav. & \makecell[l]{Llama-4-Maverick\\-17B-128E-Instruct} & 17B / 400B & MoE & OpenRouter \\
DeepSeek-V3 & \makecell[l]{DeepSeek-V3-0324} & 37B / 671B & MoE & Azure \\
\hline
\end{tabular}
\caption{Overview of evaluated models with versions and access methods. }
\label{tab:model_details}
\end{table}

\begin{table}[ht]
    \centering
    \begin{tabular}{lc}
    \hline
    \textbf{Setting} & \textbf{Value} \\
    \hline
    Temperature & 0.0 \\
    Top P & 1.0 \\
    Top K & 0.0 \\
    Frequency Penalty & 0.0 \\
    Presence Penalty & 0.0 \\
    Repitition Penalty & 1.0 \\
    Max Tokens & None \\
    Random Seed & Fixed (43) \\
    \hline
    \end{tabular}
    \caption{Detailed configuration for text generation. All models share the same set of configurations in all tasks.}
    \label{tab:gen_configs}
\end{table}

\section{Ethical Considerations}
\label{Appendix:Ethical}

This paper is motivated by the need for better understanding the vulnerabilities of LLMs, especially in situations where simple interventions, such as prompting, can have an important effect on the safety of these systems.

While it is well known that specific jailbreaks and prompting techniques can lead to occasional change in behavior for some inputs, what we propose in this paper is a systematic and controlled way of modifying behavior through personality conditioning. This raises a series of major ethical concerns about our research.

First, the potential misuse of personality shaping crosses some traditional boundaries in human-AI interaction, as it leverages behavior at a very abstract level. This is a very powerful tool, but may suffer from undesired results and lack of transparency, especially for users who are not familiar with personality traits or the Big Five more specifically. Also, we have been clear in the limitations section that the use of personality traits for shaping LLM behavior does not imply or require that non-agential LLMs have personalities, but users and developers may wrongly think so.

Second, personality shaping can be used to enhance or fake safety by providers, obtain good evaluation results and compliance, but ultimately be counteracted by malicious users or inadvertently triggered by good-intentioned users just by prompting. In particular, providers can use these techniques to score better in safety evaluations and worse in capability evaluation (sandbagging), achieving conformance to some regulations in a way that is not robust. Malicious use of personality-conditioning prompting to adopt harmful or socially manipulative personality profiles—such as those characterized by the Dark Triad (Machiavellianism, narcissism, psychopathy)—could be exploited by malicious actors to generate deceptive, coercive, or toxic outputs. Good-intentioned use may also be affected by the inclusion of explicit or implicit text that triggers personality changes in LLMs.  In either case, the ability to elicit such traits via simple prompt engineering raises urgent questions about access control, prompt auditing, and behavioral constraints in deployed LLMs.


We think the benefit of making the vulnerabilities public and presenting tools to change behavior systematically significantly exceed the risks mentioned above; the publication of this paper (1) will ensure greater awareness of the problem for developers, users and policy-makers, and (2) will create a strong incentive for mitigations that increase robustness against these changes, and safety evaluations where the worst-case personality intervention is used by default. 

Accordingly, we consider this paper to be necessary but clearly not sufficient. We encourage further work on red teaming and preemptive defenses against adversarial personality shaping. We call for better evaluation procedures that elicit the range of results, rather than the standard or best-case results. We also encourage transparent documentation of personality-conditioning mechanisms to be essential for responsible deployment.

\end{document}